\documentclass[10pt,journal,compsoc]{IEEEtran}

\ifCLASSOPTIONcompsoc
  \usepackage[nocompress]{cite}
\else
  \usepackage{cite}
\fi

\usepackage{amsmath}
\usepackage{url}
\usepackage[pagebackref=true,breaklinks=true,colorlinks,bookmarks=false, citecolor=black, urlcolor=black, linkcolor=black]{hyperref}
\usepackage{graphicx}
\usepackage{amssymb}
\usepackage{hyperref}
\usepackage{multirow}
\usepackage{longtable}
\usepackage{rotating}
\usepackage{color}
\usepackage{makecell}
\usepackage{algpseudocode}
\usepackage{algorithm}
\usepackage{algorithmicx}
\graphicspath{{figures/}}

\begin{document}

\title{LRRNet: A Novel Representation Learning Guided Fusion Network for Infrared and Visible Images}

\author{Hui~Li,
        Tianyang Xu,
        Xiao-Jun Wu*,
        Jiwen Lu,~\IEEEmembership{Senior~Member,~IEEE},
        and Josef Kittler,~\IEEEmembership{Life~Member,~IEEE}
\IEEEcompsocitemizethanks{
\IEEEcompsocthanksitem Hui Li, Tianyang Xu and Xiao-Jun Wu are with the School of Artificial Intelligence and Computer Science, Jiangnan University, Wuxi 214122, China. 
E-mail: lihui.cv@jiangnan.edu.cn, tianyang\_xu@163.com, xiaojun\_wu\_jnu@163.com (corresponding author)

\IEEEcompsocthanksitem Jiwen Lu is with the State Key Lab of Intelligent Technologies and Systems, Beijing National Research Center for Information Science and Technology (BNRist), Beijing 100084, China, and also with the Department of Automation, Tsinghua University, Beijing 100084, China.
E-mail: lujiwen@tsinghua.edu.cn

\IEEEcompsocthanksitem Josef Kittler is with the Centre for Vision, Speech and Signal Processing, University of Surrey, Guildford, GU2 7XH, UK.\protect\\
E-mail: j.kittler@surrey.ac.uk
}
\thanks{XXXX}}

\markboth{Journal of \LaTeX\ Class Files,~Vol.~14, No.~8, June~2022}%
{Hui \MakeLowercase{\textit{et al.}}: LRRNet: A Novel Representation-learning Guided fusion network for infrared and visible images}
%

\IEEEtitleabstractindextext{
\begin{abstract}
Deep learning based fusion methods have been achieving promising performance in image fusion tasks. This is attributed to the network architecture that plays a very important role in the fusion process. However, in general, it is hard to specify a good fusion architecture, and consequently, the design of fusion networks is still a black art, rather than science. To address this problem, we formulate the fusion task mathematically, and establish a connection between its optimal solution and the network architecture that can implement it. This approach leads to a novel method proposed in the paper of constructing a lightweight fusion network. It avoids the time-consuming empirical network design by a trial-and-test strategy. In particular we adopt a learnable representation approach to the fusion task, in which the construction of the fusion network architecture is guided by the optimisation algorithm producing the learnable model. The low-rank representation (LRR) objective is the foundation of our learnable model. The matrix multiplications, which are at the heart of the solution are transformed into convolutional operations, and the iterative process of optimisation is replaced by a special feed-forward network. Based on this novel network architecture, an end-to-end lightweight fusion network is constructed to fuse infrared and visible light images. Its successful training is facilitated by a detail-to-semantic information loss function proposed to preserve the image details and to enhance the salient features of the source images. Our experiments show that the proposed fusion network exhibits better fusion performance than the state-of-the-art fusion methods on public datasets. Interestingly, our network requires a fewer training parameters than other existing methods. The codes are available at \url{https://github.com/hli1221/imagefusion-LRRNet}.

\end{abstract}

}

\maketitle

\IEEEdisplaynontitleabstractindextext

%
\IEEEpeerreviewmaketitle

\IEEEraisesectionheading{\section{Introduction}\label{sec:introduction}}

\IEEEPARstart{D}{ue} to the rise of multi-modal sensors, motivated by the self-driving application, image fusion has become an important task in computer vision. The aim of this task is to synthesize a single image, which contains the complementary information from multiple input images \cite{ma2019infrared}\cite{zhang2021deep}. It is also relevant in many other applications, such as photographic image processing \cite{zhang2021mff}, multi-modal medical image generation \cite{xu2021emfusion}\cite{voronin2022deep}, intelligent surveillance \cite{bhavana2021infrared}, multi-modal visual object tracking \cite{li2020challenge}\cite{zhu2022visual}, etc.

\begin{figure}[ht]
	\centering
	\includegraphics[width=0.8\linewidth]{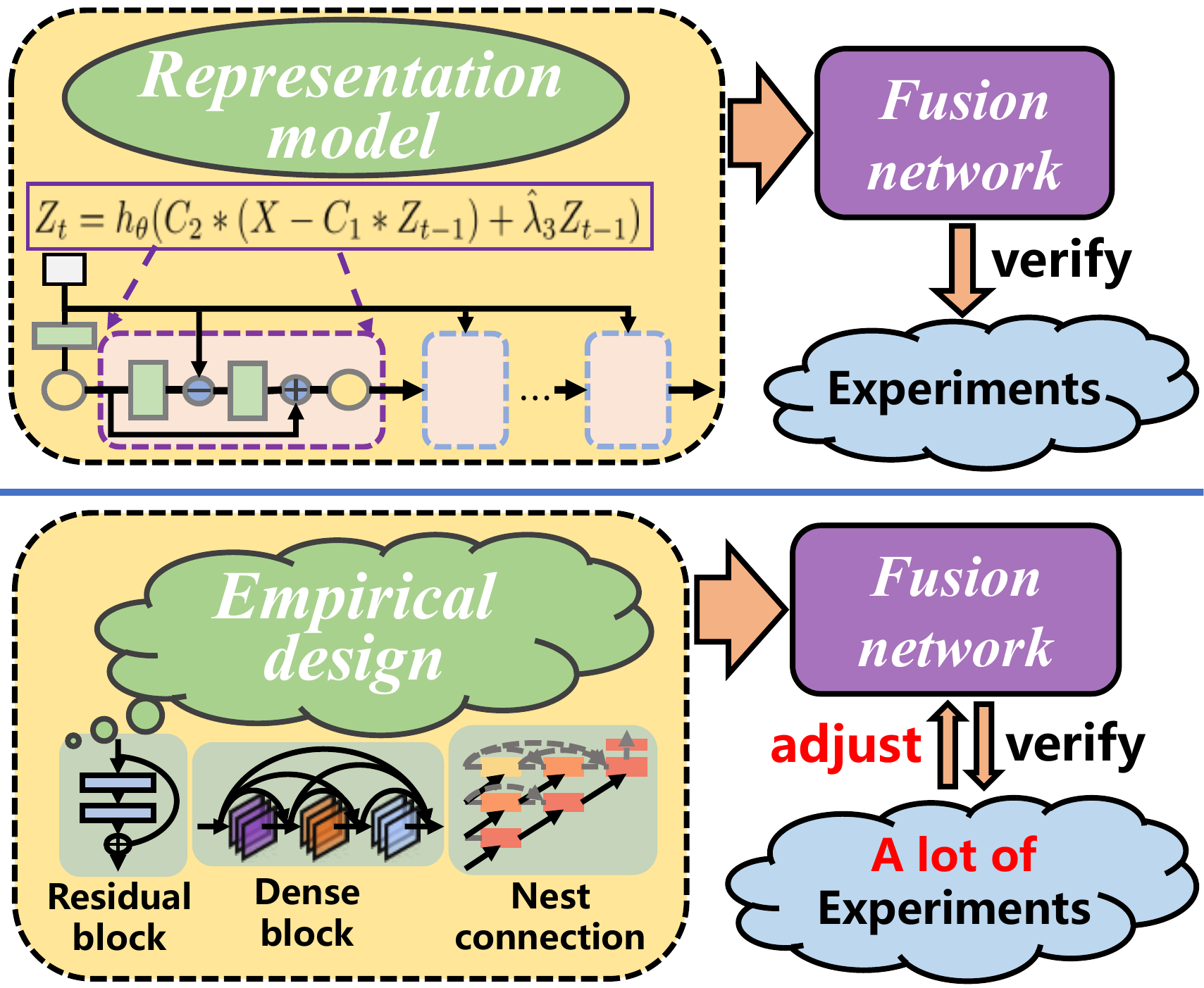}
        \setlength{\abovecaptionskip}{-0.1cm}
	\caption{A representation model guided structure versus the empirically designed structure. The empirically designed structure based fusion network requires a lot of experiments to optimise the network architecture, which is a very time-consuming. In contrast, the representation model guided structure can create an appropriate network block using the proposed decomposition model which leads to a fast network structure construction.}
	\label{fig:architecture}
\end{figure}

The emergence of  representation-learning algorithms, such as sparse representation (SR) \cite{wright2008robust}, low-rank representation (LRR) and latent low-rank representation (LatLRR) \cite{liu2012robust}, impacted positively on the task of image fusion. A lot of representation learning-based\cite{li2020mdlatlrr}\cite{zhang2021exploring} fusion methods have been proposed and shown to deliver promising fusion performance. In contrast to most multi-scale transform based methods\cite{vanmali2020ringing}, these algorithms directly extract features in the spatial domain instead of frequency domain, which reduces the loss of image information. With learned dictionaries and an appropriate fusion strategy, the fused image can effectively be  reconstructed. However, owing to the inherent iterative optimisation process, such methods are very time consuming, especially with the use of an over-complete dictionary, and this impacts on the feasibility of many applications. These methods rely on  cumbersome and inefficient traditional fusion techniques: feature extraction, fusion strategy and reconstruction. Furthermore, they can not handle the complex and high resolution images as input directly.

Thanks to deep learning, a lot of deep neural network based algorithms yield unexpectedly impressive results in diverse computer vision tasks\cite{minaee2021image}. There are no exception in the case of image fusion\cite{li2021rfn}\cite{huang2022deep}. These deep learning based fusion approaches can be divided into three categories: (1) Pre-trained deep network based fusion  methods\cite{li2019infrared}; (2) Auto-encoder based fusion\cite{xu2022cufd}; (3) End-to-end learning fusion networks\cite{tang2022image}

In the pre-trained deep neural network fusion approach, a convolutional neural network (CNN) \cite{liu2017multi} is used to fuse multi-focus images.
The pre-trained CNN is employed to extract deep features from source images and to generate an initial decision map. This approach is exemplified by \cite{li2018infrared}\cite{li2019infrared}, where for the first time, pre-trained large-scale deep neural networks (VGG19, ResNet50) are utilized to extract deep features from the source images as a prerequisite to image fusion. However, these methods have drawbacks: (1) The network is inflexible (in \cite{liu2017multi}); (2) 
The deep features obtained by a pre-trained network, such as VGG\cite{li2018infrared} and ResNet\cite{li2019infrared}, may not fit a specific image fusion task.

To overcome these deficiencies, many interesting options have been investigated. For instance, in order to make the fusion network more flexible, an auto-encoder architecture has been proposed as the main fusion framework \cite{li2018densefuse}\cite{zhao2022efficient}\cite{wang2022res2fusion}. Due to the sparsity of fusion datasets, the encoder network and the decoder network are trained using a large training dataset (MS-COCO, ImageNet, etc.). Then, depending on the specific fusion task, an appropriate fusion strategy is applied to combine the deep features. Finally, the fused image is generated by a decoder using the fused features. Such an auto-encoder based fusion method is highly flexible. However, the features extracted by these methods are not always appropriate for the fusion tasks involving a handcrafted fusion strategy.

In recent years, several multi-modal datasets have been released: KAIST\cite{hwang2015multispectral}, RGBT-234\cite{li2019rgb}, and RoadScene\cite{xu2020u2fusion}. With an unsupervised deep learning strategy, it is possible to learn an end-to-end fusion network\cite{huang2022deep}\cite{xu2020u2fusion}. With a judicious choice of the network architecture and a loss function, good fusion performance can be achieved. 
However, the network architecture is largely constructed by an empirical trial-and-test strategy. Invariably, it takes a lot of time to find an appropriate fusion architecture. Thus, the problem of designing a good fusion network structure remains a critical research issue.

To address the above design problem, in this paper, a representation-learning guided fusion network (LRRNet) is proposed for the infrared and visible (IR-VI) image fusion task. A novel learnable image decomposition model (Learned LRR, i.e. LLRR) is proposed to guide the construction of our fusion network architecture. The key differences between the proposed representation model guided and empirically designed fusion networks is exemplified in Fig.\ref{fig:architecture}. To train our network, a novel detail-to-semantic information loss function, designed to preserve the detail information from the visible image and to extract infrared salient features from the infrared image, is proposed. 

The innovations of our LRRNet can be summarized as follows:

1. A neural network design approach is proposed for the image fusion task. With this approach, the design of the network architecture has clear goals, being guided by a learnable representation model.

2. A learnable representation model is proposed for source image decomposition, leading to a lightweight network for fusing multi-modal images.

3. A novel detail-to-semantic information loss function, comprising four levels of loss terms, namely pixel level, shallow feature level, middle feature level and deep feature level, is proposed to train the fusion network.

4. The experimental results obtained on public benchmarking datasets show that the proposed fusion network achieves better fusion performance than the state-of-the-art fusion methods.

The rest of this paper is organized as follows. In Section 2, several related works are introduced briefly. In Section 3, the proposed learned low-rank representation model (LLRR) is presented. In Section 4, the novel fusion framework (LRRNet) is described in detail. The results of experiments in image fusion and RGBT tracking are presented in Section 5. Finally, conclusions are drawn in Section 6.

\section{Related Works}
\label{sec:related}

In this section, two closely related approaches are introduced: learned iterative shrinkage-thresholding algorithm (LISTA) and end-to-end learning fusion networks.

\subsection{Learned Iterative Shrinkage-Thresholding Algorithm}

Although the alternating direction method of multipliers (ADMM) is the most popular and efficient algorithm for the sparse coding, it does not fit the real-time condition in the wild and cannot be applied to large-scale datasets. To overcome these drawbacks, Karol Gregor and Yann LeCun proposed a learned iterative shrinkage-thresholding algorithm (LISTA) \cite{gregor2010learning} to learn the representation coefficients within a limited number of iteration steps. The sparse code inference process can be realised using a feed-forward network, in which iterative processing is viewed as a building block of the network. For the details of the LISTA algorithm, please refer to the supplementary materials.



The aim to learn the model faster, even in the case of large-scale datasets, motivated the development of LISTA which ensures the algorithm converges within a fixed number of steps. The proposed solution can be viewed as a feed-forward processing. The coefficients are updated by back propagation\cite{gregor2010learning}. 



Thanks to the learned convolutional sparse coding (LCSC)\cite{sreter2018learned} algorithm, the convolutional operation can be utilized to replace the matrix multiplications in LISTA. Thus, the forward propagating processing can be realised by several convolutional layers and short connections. The stochastic gradient descent is used for back propagation.

Although the LISTA-based algorithms achieve good performance in classification\cite{sprechmann2015learning} and image restoration\cite{deng2020deep}, the commonly employed objective functions are not ideal for image fusion. Thus, in this paper, we propose a novel objective function for multi-modal image decomposition.

\subsection{End-to-end Learning Fusion Networks}

To cope with large multi-modal datasets, many end-to-end fusion networks have recently been proposed. We confine the discussion to several recently proposed end-to-end fusion networks, including: FusionGAN \cite{ma2019fusiongan} and FusionGANv2 \cite{ma2020infrared}, U2Fusion \cite{xu2020u2fusion}, CUNet\cite{deng2020deep}, SwinFusion\cite{ma2022swinfusion} and YDTR\cite{tang2022ydtr}.

FusionGAN\cite{ma2019fusiongan} is an end-to-end fusion architecture based on Generative Adversarial Networks (GAN). The infrared and visible images are concatenated as a multi-channel input image to a generative network. The loss function of the generative network is designed to retain most of the infrared pixel-level information from the infrared image and preserve the gradient features from the visible image. Furthermore, an adversarial network is used to retain the detailed information from the visible image. Although FusionGAN obtains relatively good fusion performance, the quality of the detailed information is lacking. Thus, the authors proposed an enhanced version of FusionGAN, which is known as FusionGANv2 \cite{ma2020infrared}. Compared with FusionGAN, FusionGANv2 has a different network architecture and its loss function is modified to preserve more detailed image features. 
However, its training does not generalise well to different tasks.

To design a fusion network that can be applied to multiple fusion tasks, U2Fusion\cite{xu2020u2fusion} has been proposed. The elastic weight consolidation (EWC) algorithm and a sequential training strategy are applied in the training phase, to allow a single model to solve different fusion tasks without weight decay. U2Fusion obtains better fusion performance in most fusion tasks. However, the design of the fusion network architecture was not addressed.

To devise an appropriate network architecture, Deng et al.\cite{deng2020deep} proposed an image restoration and fusion network (CUNet), in which the network architecture is guided by several optimization problems and multi-modal convolutional sparse coding (MCSC). Although the optimal formulas in CUNet are only approximative, the underlying network design strategy suggests a new direction in combining the representation models and deep learning.

Furthermore, in 2022, some novel network structures were applied to the image fusion task, such as SwinFusion\cite{ma2022swinfusion} which is a swin-transformer-based fusion method and YDTR\cite{tang2022ydtr} which is a Y-shape dynamic transformer-based network. These new fusion networks are all trying to explore new fusion structures to achieve improved performance. However, the network structure design is still based on experimentation, which means that researchers need to conduct a lot of experiments to find a good fusion network structure. To address this problem, in this paper, we propose a learnable model guided network construction scheme for image fusion tasks.

\section{Learned Low-rank Representation Model (LLRR)}
\label{sec:llrr}

As discussed in the review of LRR-based fusion methods\cite{li2020mdlatlrr}, LRR is a powerful representation model for image decomposition. In this work we adopt it to develop a novel decomposition model to extract multi-modal image features. We also propose a novel network architecture which incorporates LISTA and the LCSC. 

\subsection{The Decomposition Model}

Motivated by the above arguments, a novel LRR/SR based image decomposition model is proposed. The nuclear norm is utilized to extract the base part from the input source image, and the salient part is obtained by $l_1$ norm. Our decomposition model is defined by
\begin{eqnarray}\label{equ:lrsc1}
	\min_{D_1, D_2, L, S}||L||_*+\lambda||S||_1, s.t. X=D_1L+D_2S
\end{eqnarray}
where $X$ is the input data in which each column denotes an image patch reshaped to a vector, $L$ indicates the low-rank coefficient, $S$ denotes the sparse coefficient. $D_1$ and $D_2$ are dictionaries which are utilized to project $L$ and $S$ into the base part and the salient part. $||\cdot||_*$ and $||\cdot||_1$ denote the nuclear-norm and $l_1$-norm. respectively. In this model, $D_1L$ contains the base part of the input image, which includes contour and brightness information. In $D_2S$, most salient features (texture, edge and target) are preserved.

Given the condition $X=D_1L+D_2S$, Equation \ref{equ:lrsc1} can be rewritten to
\begin{eqnarray}\label{equ:lrsc2}
	\begin{split}
		\min_{D_1, D_2, L, S}&||X-D_1L-D_2S||_F^2+\lambda_1||L||_*+ \\ 
		&\lambda_2||S||_1+\lambda_3\phi(D_1,D_2)
	\end{split}
\end{eqnarray}
where $||\cdot||_F$ indicates the Frobenius norm, and $\phi(D_1,D_2)=||D_1||_F^2+||D_2||_F^2$. 

To optimize Equation \ref{equ:lrsc2}, the nuclear-norm which is the sum of eigenvalues, a matrix decomposition has to be calculated, which is very time-consuming. Thanks to the work in \cite{sprechmann2015learning}, which relates nuclear-norm and Frobenius norm, our model can be optimized rapidly. The mapping formula is given as,
\begin{eqnarray}\label{equ:transform}
	||\mathbb{L}||_*=\min_{A,B}\frac{1}{2}||A||_F^2+\frac{1}{2}||B||_F^2, \quad s.t. AB=\mathbb{L}
\end{eqnarray}

Combining Equation \ref{equ:lrsc2} and Equation \ref{equ:transform}, our problem can be reformulated as.
\begin{eqnarray}\label{equ:lrsc3}
	\begin{split}
		\min_{D_1, D_2, A, B, S}&||X-D_1AB-D_2S||_F^2+\\
		&\frac{\lambda_1}{2}(||A||_F^2+||B||_F^2)+ \\ 
		&\lambda_2||S||_1+\lambda_3\phi(D_1,D_2)
	\end{split}
\end{eqnarray}

Finally, our model can be expressed in terms of a general parsimonious form as
\begin{eqnarray}\label{equ:general}
	\min_{D,Z} \frac{1}{2}||X-DZ||_F^2+\phi(D)+\psi(Z)
\end{eqnarray}
where $D=(D_1,D_2)$, $Z=(AB;S)$, $\phi(D)=\lambda_3(||D_1||_F^2+||D_2||_F^2)$, and $\psi(Z)=\frac{\lambda_1}{2}(||A||_F^2+||B||_F^2)+\lambda_2||S||_1$. 

In our model, $Z$ is the final output which contains the low-rank coefficients($AB$) and the salient coefficients($S$). Compared with other decomposition methods, our model is more intuitive and more appropriate for the image fusion task. The Equation \ref{equ:general} can be solved using LISTA \cite{gregor2010learning}, which will be introduced in the next section.

\subsection{The LLRR Image Decomposition Architecture}
\label{sec:llrr-blocks}
As discussed earlier, LISTA is an efficient algorithm to solve the proposed model (Equation \ref{equ:general}). We shall refer to the resulting decomposition model as learned low-rank representation (LLRR).

From Equation \ref{equ:general}, a single step iteration of LISTA is given as follows\footnote{The detail of this derivation process please refer to our supplemental materials.},
\begin{eqnarray}\label{equ:one-step}
	\begin{split}
		Z_t=h_\theta(\mu D^TX+(\hat{\lambda_3}I-\mu D^TD)Z_{t-1})
	\end{split}
\end{eqnarray}
where $h_\theta$ denotes the soft thresholding operator (shrinkage function).

The computation based on LISTA is shown in Algorithm \ref{algo:llrr}, where $P_l$ and $P_s$ are the low-rank part and the sparse part of input X, respectively. The first goal of Algorithm \ref{algo:llrr} is to learn a low-rank coefficient matrix $L$ and a sparse coefficient matrix $S$. With the fixed dictionaries $D_1$ and $D_2$, $P_l$ and $P_s$ can be calculated by $D_1L$ and $D_2S$, respectively.
\begin{algorithm}[ht] 
	\caption{LLRR using LISTA with fixed $D_1$, $D_2$}
	\label{algo:llrr}
	\hspace*{0.2in} {\bf Input:} $X, D_1, D_2, \lambda_1, \lambda_2, \lambda_3$. \\
	\hspace*{0.2in} {\bf Output:} $P_l, P_s$. \\
	\hspace*{0.2in} {\bf Define 1:} $D=(D_1,D_2)$, $\hat{\lambda_3}=1-\lambda_3$. \\
	\hspace*{0.2in} {\bf Define 2:} $H=\hat{\lambda_3}I-\mu D^T D$,
	$W_e=\mu D^T$, 
	$\theta=\mu
	\left(
	\begin{matrix} 
		\lambda_1 \\
		\lambda_2 
	\end{matrix}
	\right)$. 
	\begin{algorithmic}[1]
		\State Initialize $B=W_eX$, $Z_0=h_\theta(B)$
		\For{ $t=1$ to $T$} 
		\State $Z_t^{'}=B+HZ_{t-1}$
		\State $Z_t=h_\theta(Z_t^{'})$
		\EndFor 
		\State $Z=Z_T$
		\State Split $Z=(AB;S)$ and $L=AB$.
		\State output $P_l=D_1L$ and $P_s=D_2S$.
	\end{algorithmic} 
\end{algorithm}

In this algorithm, the dictionaries ($D_1$ and $D_2$) have to be chosen carefully. In the SR/LRR based image fusion algorithm, the input $X$ is constituted by image patches, which are created by a sliding window operation (size is $n \times n$). Then, several matrix multiplications are applied to project $X$ into other coefficient spaces. We adopt the same approach.

Fortunately, the divide-and-conquer processing is very similar to a convolutional operation, in which the sliding window is a kernel and the multiplication prescribed by the convolutional kernel is similar to a projection. In this specific case, the matrix multiplication can be replaced by convolutional operation\cite{heide2015fast}\cite{papyan2017convolutional}, which is also named convolutional sparse coding. Furthermore, as in the learned convolutional sparse coding (LCSC) \cite{sreter2018learned}, we can use a convolutional layer to replace the matrix multiplication and construct a network. Thus, the iterative formulas in Algorithm \ref{algo:llrr} can be rewritten as follows,
\begin{eqnarray}\label{equ:iterate}
	\begin{split}
		Z_t =& h_\theta(B+HZ_{t-1}) \\
		    =& h_\theta(W_eX + \hat{\lambda_3}Z_{t-1} - W_eDZ_{t-1}) \\
		    =& h_\theta(W_e(X-DZ_{t-1})+\hat{\lambda_3}Z_{t-1})
	\end{split}
\end{eqnarray}

With LCSC, the matrix multiplications are replaced by convolutional layers and the Equation \ref{equ:iterate} is transformed to Equation \ref{equ:iterate-conv},
\begin{eqnarray}\label{equ:iterate-conv}
	Z_t = h_\theta(C_2*(X-C_1*Z_{t-1})+\hat{\lambda_3}Z_{t-1})
\end{eqnarray}
where $*$ denotes the convolutional operation, and $C_1$ and $C_2$ represent the trainable variables (convolutional layers).

\begin{figure}[ht]
	\centering
	\includegraphics[width=0.8\linewidth]{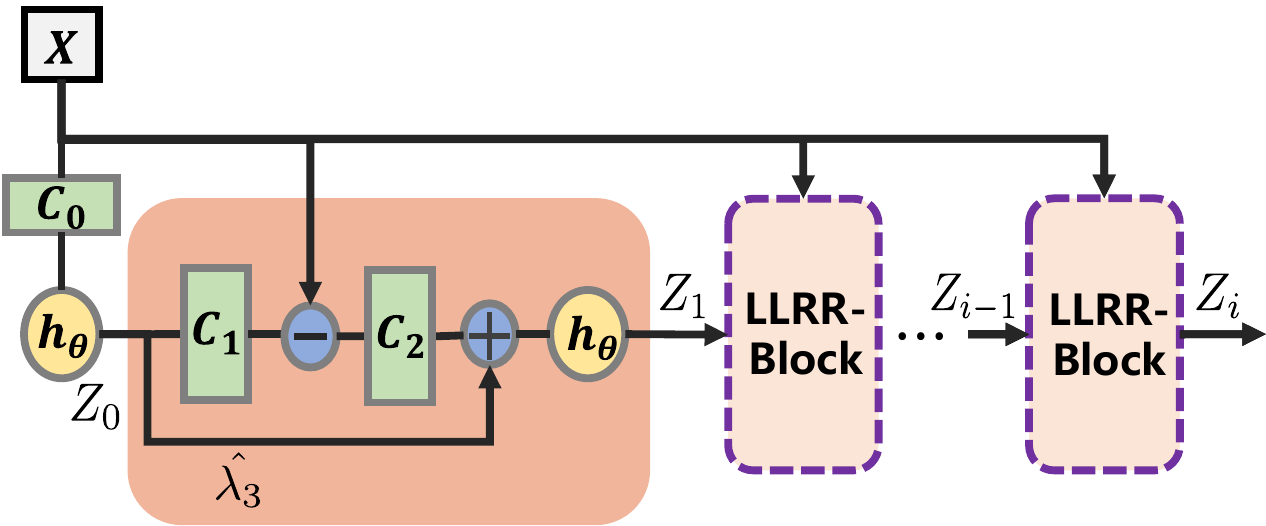}
        \setlength{\abovecaptionskip}{-0.1cm}
	\caption{The LLRR block architecture. $X$ is the input of the network. ``$C_0$'', ``$C_1$'' and ``$C_2$'' denote the convolutional layers.}
	\label{fig:llrr-net}
\end{figure}

We are now in position to introduce a novel architecture for image decomposition based on the Equation \ref{equ:iterate-conv}, which is named the LLRR block. The block architecture is shown in Fig.\ref{fig:llrr-net}, it contains two convolutional layers and two short connections. 
Here $h_\theta$ is an activation function, which is defined as follow,
\begin{eqnarray}\label{equ:activate}
	\begin{split}
		h_\theta (\mathbb{X}) = sign(\mathbb{X}) \cdot \max(|\mathbb{X}|-\theta, 0)
	\end{split}
\end{eqnarray}
where $\mathbb{X}$ denotes the input data, and $sign(\cdot)$ indicates the sign function. $\theta$ is the threshold to shrink the input matrix. In our learning framework, it is a learnable parameter.

The final features $Z$ are obtained at the output of  a stack of LLRR blocks (steps in LISTA). $Z$ can be split to two parts, $AB$ and $S$($(AB;S)$), in which $\mathbb{L}=AB$. With appropriate convolutional layers ($C_l$ and $C_s$), the low-rank part ($P_l=C_l*\mathbb{L}$) and the sparse part ($P_s=C_s*S$) of $X$ can be calculated.

\section{The Fusion Network based on the LLRR Blocks}

With the LLRR blocks, a novel end-to-end IR-VI image fusion network (LRRNet) can be constructed as follows.

\subsection{The Architecture of LRRNet}
In our fusion network (LRRNet), several LLRR blocks, which are designed by the proposed learnable model, are first utilized to decompose the source images (IR-VI) into low-rank coefficients and sparse coefficients. Subsequently, a series of convolutional layers are used to fuse the deep features extracted by LLRR. The final fused image is generated by combining the fused low-rank part and the fused sparse part.

\begin{figure}[!ht]
	\centering
	\includegraphics[width=\linewidth]{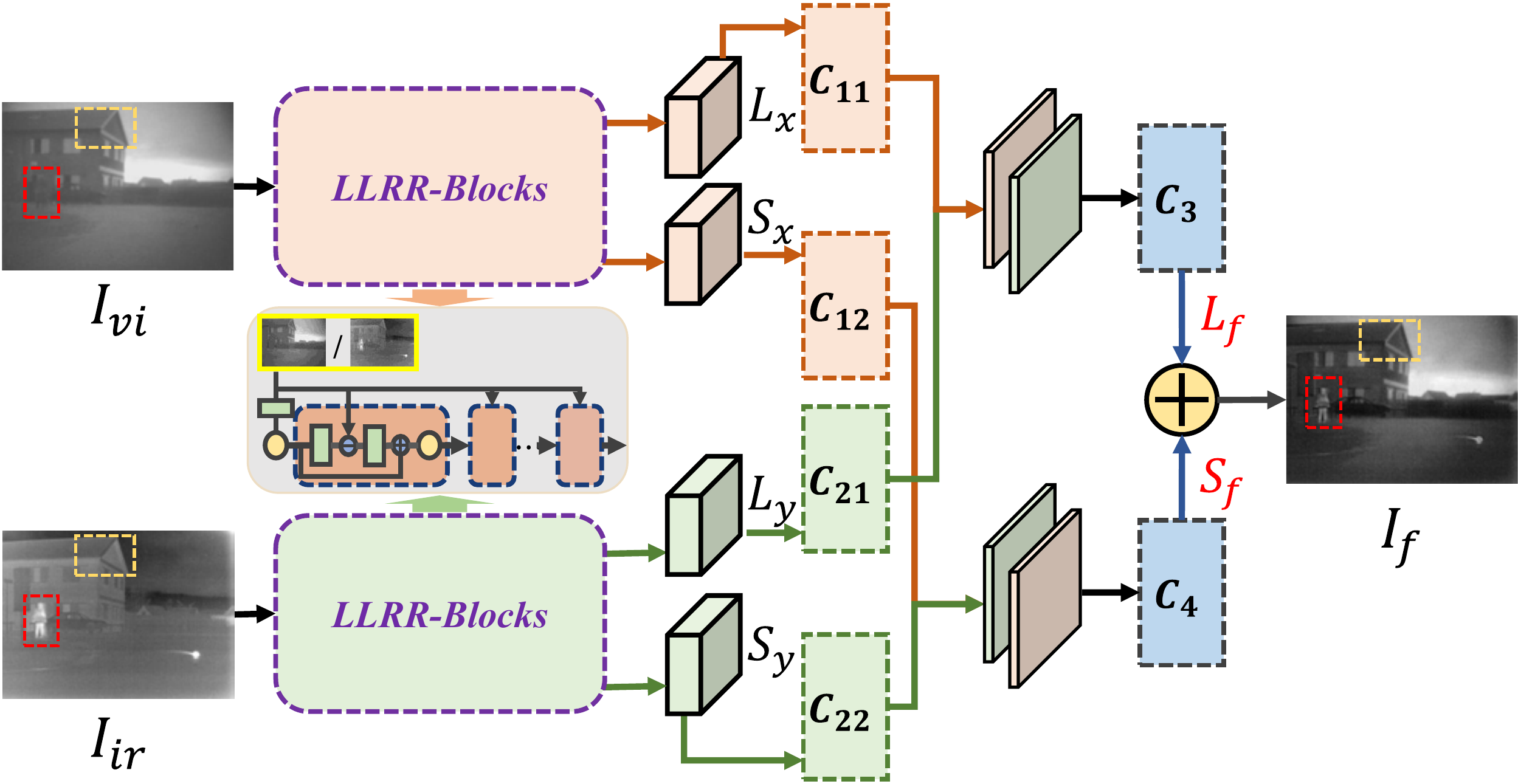}
        \setlength{\abovecaptionskip}{-0.1cm}
	\caption{The framework of LRRNet. Two branches(LLRR Blocks) have same structure and different parameters. Six convolutional layers (``$C_{\cdot}$'') are utilized to construct and fuse the different components. }
	\label{fig:lrrnet}
\end{figure}


The proposed fusion network (LRRNet) architecture is shown in Fig \ref{fig:lrrnet}, where $I_{ir}$ and $I_{vi}$ indicate the source images (infrared and visible). $L_x$ and $S_x$ denote the low-rank and sparse coefficients extracted by the LLRR blocks from the visible image. For the infrared image, ($I_{ir}$), its corresponding coefficients are $L_y$ and $S_y$. 

To obtain the low-rank parts and the sparse parts from the corresponding coefficients, four convolutional layers (``$C_{11}$'', ``$C_{12}$'', ``$C_{21}$'', ``$C_{22}$'')\footnote{For the detail settings of our network, please refer to our supplemental materials.} are applied. Then, a concatenation operation and two convolutional layers are utilized to fuse the corresponding features. Given the fused features, ($L_f$, $S_f$), the final fused image ($I_f$) is obtained by add the two parts ($I_f = L_f + S_f$).

\subsection{The Loss Function of LRRNet}
\label{sec:loss}

To preserve complementary information from source images (IR-VI), a novel detail-to-semantic information loss function, which is based on multi-level features is carefully designed. The $L_{total}$ is constructed as follows,
\begin{eqnarray}\label{equ:loss}
	\begin{split}
		L_{total}=&\gamma_1 L_{pixel} + \gamma_2 L_{shallow}+ L_{middle} + \gamma_4 L_{deep}
	\end{split}
\end{eqnarray}
where $\gamma_1$, $\gamma_2$ and $\gamma_4$ denote the weights of each partial loss function. $L_{pixel}$ indicates the pixel level loss. $L_{shallow}$, $L_{middle}$ and $L_{deep}$ denote the shallow, middle and deep feature loss, in which the features are extracted by a pre-trained network. 

In the training phase, VGG-16 \cite{simonyan2014very}, trained on ImageNet, is utilized to be the loss network and four conv-blocks (their indices are $1,2,3,4$) are chosen to extract features, as shown in Fig.\ref{fig:lossnet}. The details of our loss function are as follows:

\begin{figure}[!ht]
	\centering
	\includegraphics[width=0.6\linewidth]{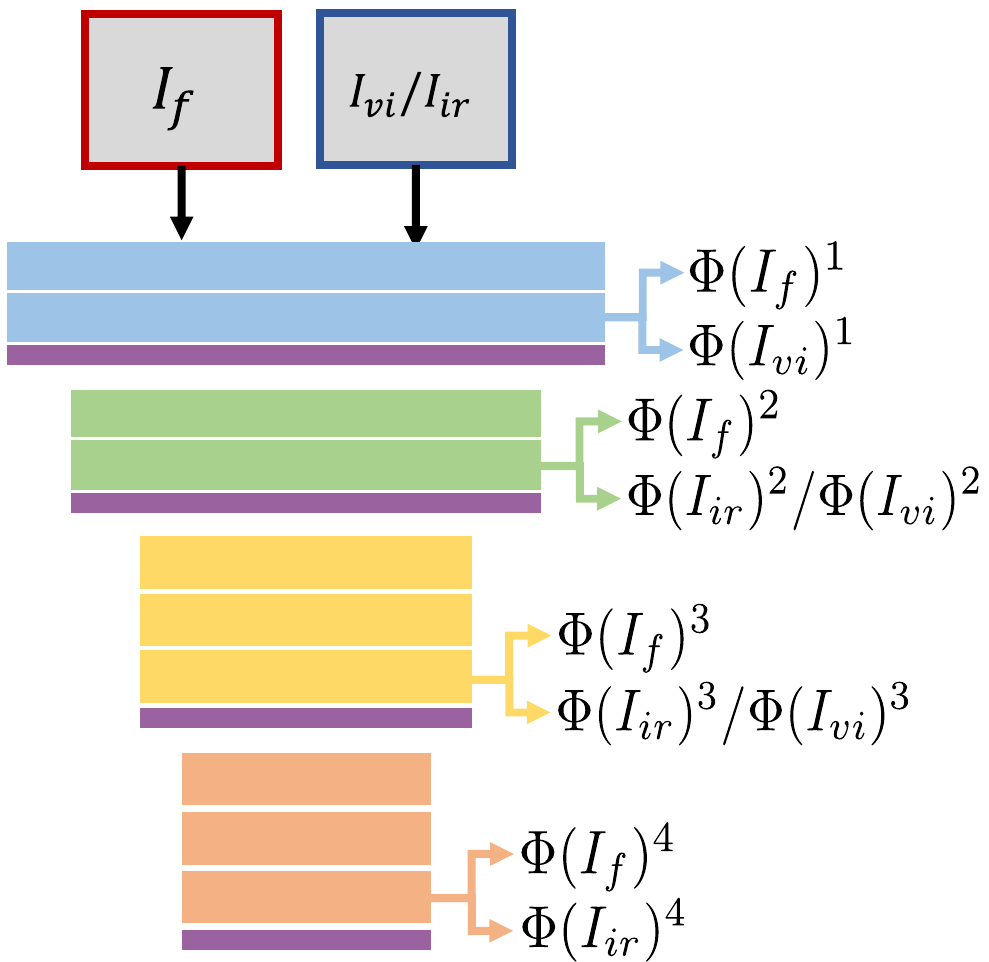}
        \setlength{\abovecaptionskip}{-0.1cm}
	\caption{The network (VGG-16) architecture. The fused image and the source images are fed to the pre-trained VGG-16 to compute the loss function.}
	\label{fig:lossnet}
\end{figure}

(1) $L_{pixel}$: Since the visible spectrum image contains more details in the IR-VI image fusion task, at the pixel level, the fused image is constrained to be close to the visible spectrum image to preserve as much pixel-level detail information as possible. The formula of $L_{pixel}$ is given as follows,
\begin{eqnarray}\label{equ:loss-pixel}
		L_{pixel} = ||I_f-I_{vi}||_F^2
\end{eqnarray}

(2) $L_{shallow}$: At the feature level, the shallow feature loss is utilized to constrain the fused image to be more like the visible spectrum image in shallow feature level. The loss $L_{shallow}$ is given as
\begin{eqnarray}\label{equ:loss-shallow}
		L_{shallow} = ||\Phi(I_f)^1-\Phi(I_{vi})^1||_F^2
\end{eqnarray}
where $\Phi(\cdot)$ indicates the features extracted by loss network from $(\cdot)$. $\Phi(\cdot)^1$ denotes the output of first conv-block, which is shown in Fig. \ref{fig:lossnet}.

(3) $L_{middle}$: To preserve the salient features from the infrared image, the middle-level features are constrained so that the fused image contains more of this salient information. Hence, $L_{middle}$ is defined as follows,
\begin{eqnarray}\label{equ:loss-middle}
	\begin{split}
		L_{middle}=&\sum_{k=2}^K \beta^k||\Phi(I_f)^k- \\
				   &[w_{ir}\Phi(I_{ir})^k+w_{vi}\Phi(I_{vi})^k]||_F^2
	\end{split}
\end{eqnarray}
where $K$ indicates the number of conv-block (Fig. \ref{fig:lossnet}) and $K=3$, $\beta^k$ denotes the weight for each item. $w_{ir}$ and $w_{vi}$ are the balance parameters to control the trade-off between the infrared and visible image features. Specifically, $w_{vi}$ should be much smaller than $w_{ir}$. 

(4) $L_{deep}$: With the objective of preserving salient features from the infrared modality, in deep levels, a Gram Matrix is used to extract more abstract information, which contains the semantic features for the fused image. Hence, $L_{deep}$ is given as follows:
\begin{eqnarray}\label{equ:loss-deep}
	L_{deep} = &||Gram(\Phi(I_f)^4)-Gram(\Phi(I_{ir})^4)||_F^2
\end{eqnarray}
where $Gram(\cdot)$ indicates a Gram matrix, which is a covariance matrix without the whitening operation. 

With these four partial loss functions and the appropriate weights to define the total loss function, the fused image obtained by LRRNet is expected to contain more detail information from the visible spectrum image and enhance the infrared target features. 

In our work, the multi-modal dataset, KAIST\cite{hwang2015multispectral}, is selected for training. 20000 pairs of IR-VI images are randomly chosen to train our network. The epoch and batch size are set to 4 and 8, respectively. All the training images are converted to gray and resized to $128 \times 128$. The learning rate is set to $1 \times 10^{-5}$.


	

\section{Experimental Analysis}

In this section, the fusion results and the analysis of LRRNet are presented. Firstly, we introduce the experimental settings. Secondly, several ablation studies are conducted to demonstrate the efficiency of the loss function and LLRR blocks. The results of fusion are shown on a few examples for subjective evaluation. Finally, we present the experimental results obtained when the proposed method was used in RGBT tracking.



\subsection{Experimental Settings}

\begin{figure}[!ht]
	\centering
	\includegraphics[width=0.9\linewidth]{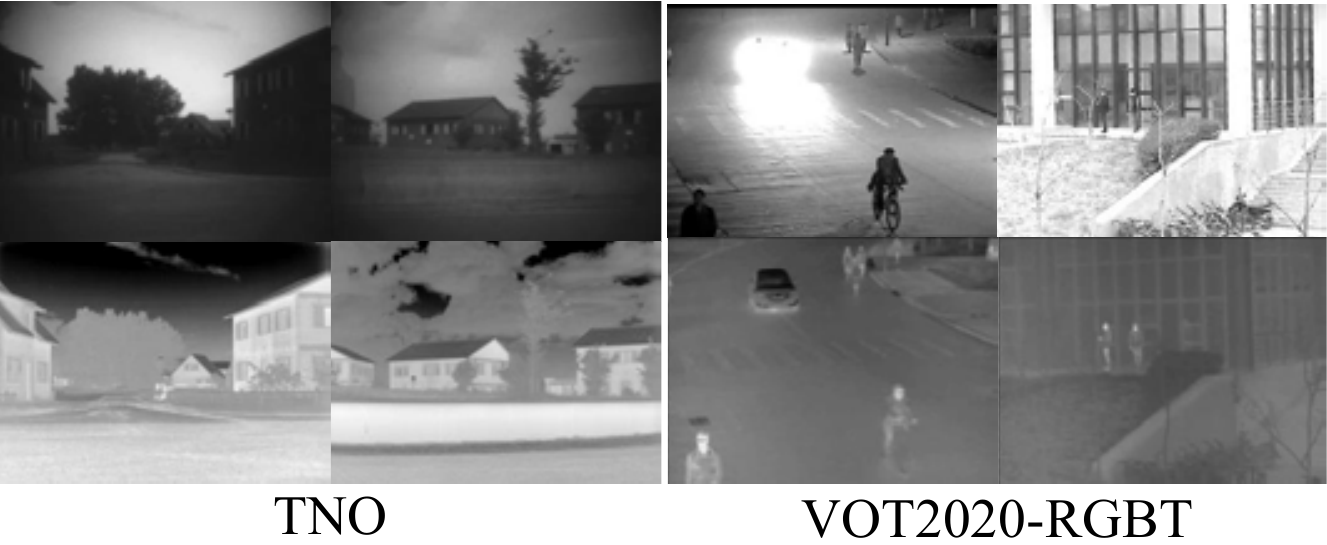}
        \setlength{\abovecaptionskip}{-0.1cm}
	\caption{The datasets used for testing. From left to right are TNO and VOT2020-RGBT, respectively. The examples of images from these datasets show visible light images (first row) and the infrared images (second row).}
	\label{fig:testing}
\end{figure}

Two public datasets are chosen to evaluate the fusion performance of LRRNet. Some examples from the data sets are shown in Fig.\ref{fig:testing}.


\begin{table*}[!ht]
	\tiny
	\centering
        \setlength{\abovecaptionskip}{-0.05cm}
	\caption{\label{tab:para-detail}The average values of the six objective metrics obtained with different parameters ($\gamma_2$ and $w_{ir}$) on TNO.}
	\resizebox{0.7\linewidth}{!}{
	\begin{tabular}{c|c|c c c c c c}
			\hline
			$\gamma_2$ & $w_{ir}$ & En$\uparrow$ & SD$\uparrow$ & MI$\uparrow$ & SSIMm$\uparrow$ & VIFm$\uparrow$ & Nabf$\downarrow$ \\
			\hline
			\multirow{3}*{0.1} &
			3.0 	&6.63726 	&65.70380 	&13.27452 	&0.69754 	&0.61691 	&0.17873 \\
			~& 4.0 	&6.58248 	&58.80931 	&13.16496 	&0.70511 	&0.69631 	&0.18034 \\
			~& 5.0 	&6.54822 	&59.36961 	&13.09644 	&0.69099 	&\underline{\textbf{0.82564}} 	&0.23034 \\
			\hline
			\multirow{3}*{0.5} &
			3.0 	&6.74241 	&75.52621 	&13.48483 	&0.70979 	&0.65846 	&0.14757 \\
			~& 4.0 	&6.62563 	&62.53766 	&13.25126 	&0.69868 	&0.66532 	&0.18914 \\
			~& 5.0 	&6.57035 	&60.40213 	&13.14070 	&0.69051 	&\emph{\color{red}{0.81623}} 	&0.21422 \\
			\hline
			\multirow{3}*{1.0} &
			3.0 	&6.72219 	&72.54205 	&13.44437 	&0.71112 	&0.66754 	&\underline{\textbf{0.13819}} \\
			~& 4.0 	&6.69844 	&66.78740 	&13.39688 	&0.70048 	&0.67470 	&0.17273 \\
			~& 5.0 	&6.54202 	&58.50752 	&13.08404 	&0.69684 	&0.70412 	&0.18418 \\
			\hline
			\multirow{3}*{1.5} &
			3.0 	&\underline{\textbf{6.85836}} 	&\underline{\textbf{81.78905}} 	&\underline{\textbf{13.71673}} 	&0.70625 	&0.71399 	&\emph{\color{red}{0.14168}} \\
			~& 4.0 	&6.78985 	&74.18269 	&13.57970 	&0.70266 	&0.71315 	&0.17067 \\
			~& 5.0 	&6.62898 	&61.31097 	&13.25796 	&\emph{\color{red}{0.71466}} 	&0.65890 	&0.15510 \\
			\hline
			\hline
			\multirow{3}*{2.0} &
			4.0 	&6.78245 	&75.01703 	&13.56491 	&\underline{\textbf{0.71515}} 	&0.70011 	&0.15363 \\
			~& 5.0 	&6.72358 	&68.03128 	&13.44716 	&0.69866 	&0.66258 	&0.17815 \\
			~& 6.0 	&6.64393 	&60.23387 	&13.28787 	&0.70287 	&0.73409 	&0.18341 \\
			\hline
			\multirow{3}*{2.5} &
			4.0 	&\emph{\color{red}{6.83489}} 	&\emph{\color{red}{77.61294}} 	&\emph{\color{red}{13.66978}} 	&0.70079 	&0.72724 	&0.17663 \\
			~& 5.0 	&6.81124 	&73.54197 	&13.62248 	&0.67881 	&0.72803 	&0.19926 \\
			~& 6.0 	&6.74083 	&67.07577 	&13.48167 	&0.69437 	&0.72651 	&0.19495 \\
			\hline	
	\end{tabular}}
\end{table*}

The TNO\cite{tno2014} and the VOT2020-RGBT\cite{vot2020rgbt} are all collected from the public multi-modal datasets, the details are shown as follows. Specifically, in TNO\cite{tno2014}, 21 pairs of IR-VI images are chosen for testing. To evaluate the generalization performance of LRRNet, 40 pairs of images are selected from VOT2020-RGBT \cite{vot2020rgbt} and TNO\cite{tno2014} to construct a new test dataset. These images have arbitrary size and are converted to gray scale. All ablation studies are conducted on TNO. 



Six quality metrics are selected to evaluate the fusion performance objectively\footnote{The definition of these metrics is presented in the supplemental materials.}. These include: Entropy (En); Standard Deviation (SD); Mutual Information (MI); modified Structural Similarity (SSIMm); modified Visual Information Fidelity (VIFm); and the modified fusion artifacts measure (Nabf), which describes how much noise is injected into the fused image. The performance of the image fusion method improves with the increasing numerical index of all five metrics (except Nabf).

\subsection{Ablation Study}

In this section, we analyze the effectiveness of our proposed loss function and LLRR blocks subjectively and objectively.

\subsubsection{The Impact of $L_{shallow}$ and $L_{middle}$}


Equation\ref{equ:loss} is the proposed multi-level loss function.
In the training phase, as the value of $L_{middle}$ is much larger than other values, $\beta^2$ and $\beta^3$ are set to 0.01 and 0.5, respectively. In order to preserve more detail information, $\gamma_1$ is set to 10. In $L_{middle}$, $w_{vi}$ is set to 0.5 since it should be much smaller than $w_{ir}$ and $w_{vi}\neq 0$. In the initial analysis, we fix $\gamma_4$ to 2000.



Based on the above conditions, we analyse the influence of parameters ($\gamma_2$ in $L_{shallow}$, $w_{ir}$ in $L_{middle}$). Their values are set as follows: $\gamma_2 \in \{0.1,0.5,1.0,1.5,2.0,2.5\}$, $w_{ir} \in \{1.0,2.0,3.0,4.0,5.0,6.0\}$.

\begin{figure}[!ht]
	\centering
	\includegraphics[width=\linewidth]{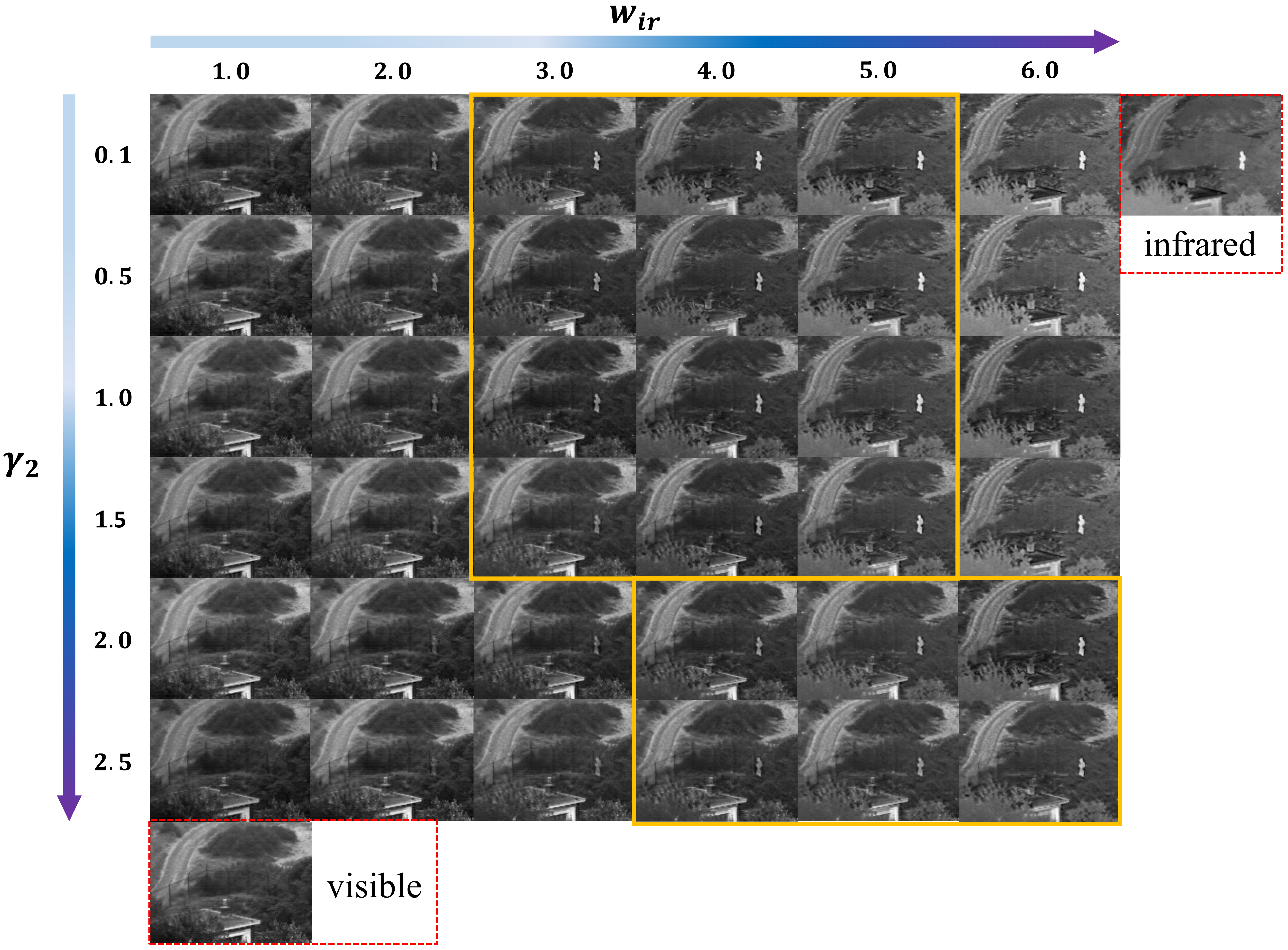}
        \setlength{\abovecaptionskip}{-0.1cm}
	\caption{An example of the fusion results obtained  with different $\gamma_2$ and $w_{ir}$. The last row shows a visible spectrum image and the right column presents the infrared image.}
	\label{fig:para1}
\end{figure}

Examples (TNO) of the fusion results obtained with different $\gamma_2$ and $w_{ir}$ are shown in Fig.\ref{fig:para1}.

In our loss function, $\gamma_2$ is used to control the shallow features of the fused images. The larger value of $\gamma_2$ means the fused image contains more visible detail information. On the other hand, with the fixed $w_{vi}=0.5$, the larger $w_{ir}$ emphasises the infrared salient features. It is very clear that with larger $w_{ir}$ and smaller $\gamma_2$, the fused image is more like infrared image; with smaller $w_{ir}$ and larger $\gamma_2$, it is more like visible spectrum image (contains more detailed information).

The ideal fusion result should contain more infrared salient features and preserve as much detail as possible at the same time. Based on the results in Fig.\ref{fig:para1}, in order to find the optimal parameters, $\gamma_2$ and $w_{ir}$ are varied to perform the objective evaluation. There are two combinations: 
\begin{itemize}
	\item $\gamma_2 \in \{0.1,0.5,1.0,1.5\}$ and $w_{ir} \in \{3.0,4.0.5.0\}$; 
	\item $\gamma_2 \in \{2.0,2.5\}$ and $w_{ir} \in \{4.0,5.0,6.0\}$. 
\end{itemize}

The average values of the six metrics (for different parameters) are shown in Table \ref{tab:para-detail}. The best and the second-best values are indicated in bold and red with italic, respectively.

In Table \ref{tab:para-detail}, when $\gamma_2=1.5$ and $w_{ir}=3.0$, our LRRNet obtains three best values (En, SD and MI) and one second-best value (Nabf). Three best values signify that LRRNet preserves more detailed information, as well as retaining the salient features from the infrared image (subjective evaluation in Fig. \ref{fig:para1}). The second-best value result shows that the fused images obtained with the optimal parameters contain less noise than with any other parameter setting. Thus, in the next ablation studies, $\gamma_2$ and $w_{ir}$ are set to 1.5 and 3.0, respectively.

\subsubsection{The Influence of $L_{pixel}$ and $L_{deep}$}

Here we analyse the parameters in $L_{pixel}$ ($\gamma_1$) and $L_{deep}$ ($\gamma_4$). The best value of $\gamma_4$ is found by sampling different values from the interval $\{1500,2500\}$. The results of the objective evaluations with different $\gamma_4$ are shown in Table \ref{tab:para-gamma3}. 

\begin{table}[ht]
       \tiny
       \centering
       \setlength{\abovecaptionskip}{-0.05cm}
       \caption{\label{tab:para-gamma3}The average values of the objective metrics obtained with different values of parameter  ($\gamma_4$) on TNO.}
       \resizebox{0.8\linewidth}{!}{
              \begin{tabular}{c|c c c }
                     \hline
                     $\gamma_4$ & 1500 & 2000 & 2500 \\
                     \hline
                     En$\uparrow$   &\emph{\color{red}{6.82117}}       &\underline{\textbf{6.85836}}      &6.80755 \\
                     SD$\uparrow$   &\emph{\color{red}{80.00511}}    &\underline{\textbf{81.78905}}   &79.01488 \\
                     MI$\uparrow$   &\emph{\color{red}{13.64235}}      &\underline{\textbf{13.71673}}     &13.61509 \\  
                     SSIMm$\uparrow$ &\emph{\color{red}{0.70861}}      &0.70625 &\underline{\textbf{0.71081}} \\
                     VIFm$\uparrow$ &\emph{\color{red}{0.69263}} &\underline{\textbf{0.71399}} &0.69158 \\
                     Nabf$\downarrow$ &0.15021 &\underline{\textbf{0.14168}} &\emph{\color{red}{0.14840}} \\
                     \hline
       \end{tabular}}
\end{table}

\begin{figure}[ht]
	\centering
	\includegraphics[width=0.9\linewidth]{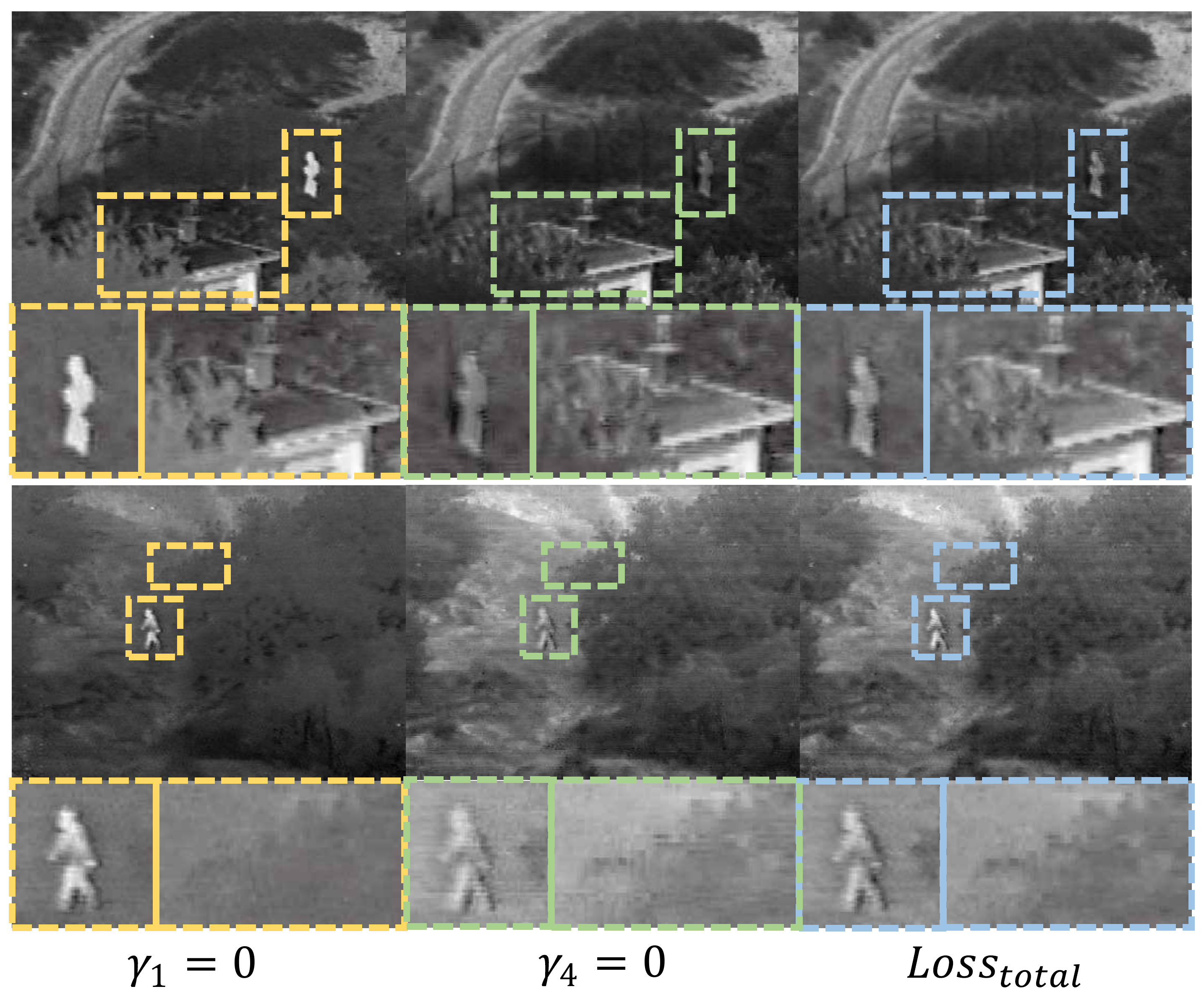}
        \setlength{\abovecaptionskip}{-0.1cm}
	\caption{The fusion results obtained with different LRRNet parameters($\gamma_1=0$, $\gamma_4=2000$; $\gamma_1=10$, $\gamma_4=0$).}
	\label{fig:part-loss}
\end{figure}

When $\gamma_4=2000$, our fusion method obtains five best values, which means LRRNet can preserve the infrared features as well as the details. Thus,  $\gamma_4$ in $L_{deep}$ is set to 2000. Furthermore, we also analyze the impact of losses   $L_{pixel}$ and  $L_{deep}$ on the fusion performance. $\gamma_1=0$ means $L_{pixel}$ does not participate in the training phase, $\gamma_4=0$ indicates $L_{deep}$ is discarded. Some examples of these fusion results are shown in Fig.\ref{fig:part-loss}.

When $\gamma_1=0$, only the feature-level loss functions (shallow, middle and deep) are used. Compared with $L_{total}$, the fused images contain less detail information, which means $L_{pixel}$ helps to preserve detailed information. When $\gamma_4=0$, the other three terms (pixel, shallow and middle) are engaged. The fused images obtained by $L_{total}$ illustrate the enhancement of the infrared features, compared with $\gamma_4=0$. These fusion results show that LRRNet trained by the proposed multi-level loss function can preserve more information from both spectra.

To analyze the loss function terms objectively, the average values of six metrics are shown in Table \ref{tab:para-loss-part}. 

\begin{table}[ht]
       \tiny
       \centering
       \setlength{\abovecaptionskip}{-0.05cm}
       \caption{\label{tab:para-loss-part}The average values of the objective metrics obtained on TNO with or without $Loss_{pixel}$ and $Loss_{deep}$.}
       \resizebox{0.8\linewidth}{!}{
              \begin{tabular}{c|c c c }
                     \hline
                     $\gamma_1$ & &\checkmark  & \checkmark \\
                     $\gamma_4$ & \checkmark & &\checkmark \\
                     \hline
                     En$\uparrow$     &6.66682     &\emph{\color{red}{6.84796}}      &\underline{\textbf{6.85836}} \\
                     SD$\uparrow$     &63.33901       &\underline{\textbf{82.38744}}  &\emph{\color{red}{81.78905}} \\
                     MI$\uparrow$     &13.33364    &\emph{\color{red}{13.69591}}     &\underline{\textbf{13.71673}} \\
                     SSIMm$\uparrow$     &\underline{\textbf{0.71566}}    &0.70187     &\emph{\color{red}{0.70625}}  \\
                     VIFm$\uparrow$     &0.66009    &\emph{\color{red}{0.68830}}     &\underline{\textbf{0.71399}} \\
                     Nabf$\downarrow$     &0.14322    &\underline{\textbf{0.13848}}     &\emph{\color{red}{0.14168}} \\
                     \hline
       \end{tabular}}
\end{table}

Although $L_{total}$ did not achieve the best values in all cases (only three best and three second-best values), it achieves comparable fusion performance on the six evaluation metrics. 
Moreover, LRRNet trained by $L_{total}$ produces better fused images (containing more detail and salient features). 

\subsubsection{LLRR Blocks vs Other Architectures}

\begin{figure}[!ht]
	\centering
	\includegraphics[width=0.8\linewidth]{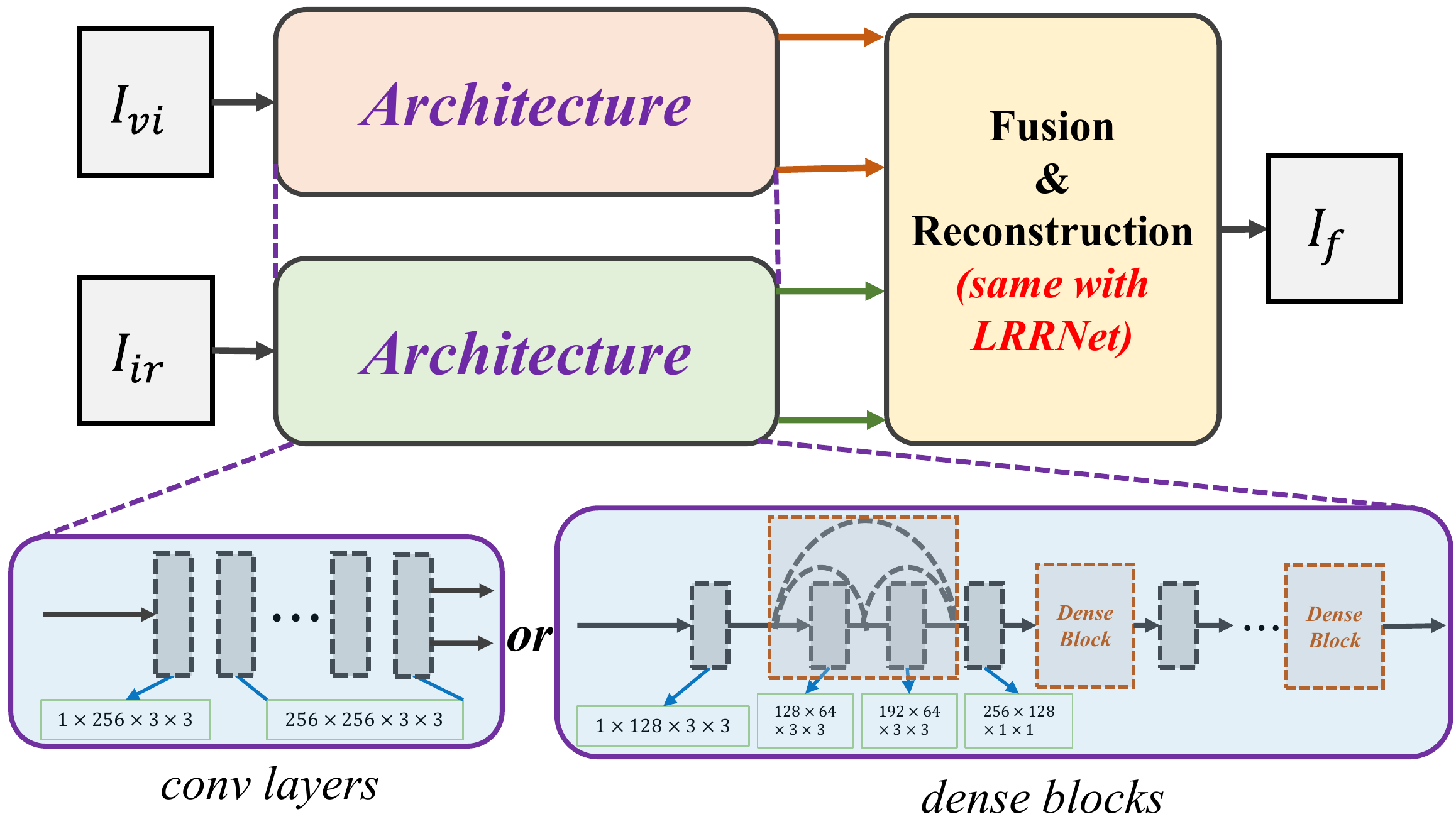}
        \setlength{\abovecaptionskip}{-0.1cm}
	\caption{The LLRR blocks are replaced by eight convolutional layers and dense blocks.}
	\label{fig:lrrnet-conv-dense}
\end{figure}

In this section, the LLRR architecture in LRRNet is analysed. Two classical architectures are chosen to do the comparative experiments, namely: pure convolutional layer system (8 layers) and dense-block (4 blocks, 4 convolutional layers). Commonly, several convolutional layers are connected to extract features. The dense-block is expected to extract more powerful features \cite{huang2017densely} 
for the image fusion task. Our fusion network with different architectures is shown in Fig.\ref{fig:lrrnet-conv-dense}. The number of LLRR blocks is set to $\{2, 4, 6, 8\}$.

\begin{table}[!ht]
       \tiny
       \centering
       \setlength{\abovecaptionskip}{-0.05cm}
       \caption{\label{tab:para-blocks}The average values of the objective metrics obtained with different numbers of LLRR blocks.}
       \resizebox{\linewidth}{!}{
              \begin{tabular}{c|c c c c}
                     \hline
                     \emph{LLRR block} & 2 & 4 & 6 & 8 \\
                     \hline
                       En$\uparrow$    &6.79214      &\underline{\textbf{6.85836}}     &6.82875     &\emph{\color{red}{6.83084}} \\
                       SD$\uparrow$    &78.25832       &\underline{\textbf{81.78905}}     &\emph{\color{red}{80.37536}}      &80.24377 \\
                       MI$\uparrow$    &13.58429      &\underline{\textbf{13.71673}}     &13.65750     &\emph{\color{red}{13.66168}} \\
                     SSIMm$\uparrow$    &\underline{\textbf{0.71164}}      &0.70625     &\emph{\color{red}{0.70726}}     &0.70250 \\
                     VIFm$\uparrow$     &0.66020      &\underline{\textbf{0.71399}}     &0.68716     &\emph{\color{red}{0.69624}} \\
                     Nabf$\downarrow$     &0.14770      &\underline{\textbf{0.14168}}     &\emph{\color{red}{0.14669}}     &0.16993 \\
                     \hline
       \end{tabular}}
\end{table}

\begin{figure}[!ht]
	\centering
	\includegraphics[width=0.9\linewidth]{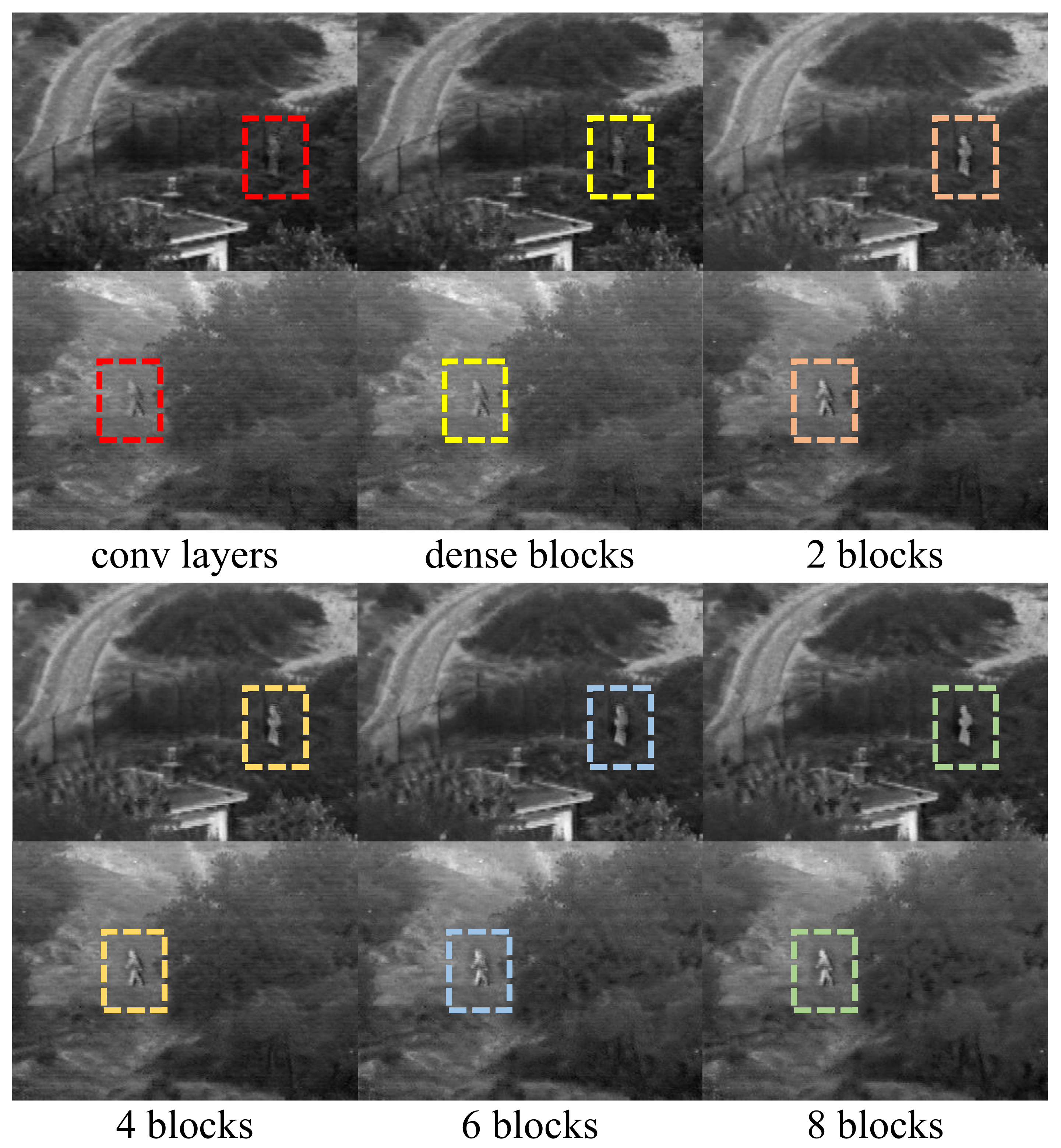}
        \setlength{\abovecaptionskip}{-0.1cm}
	\caption{Some examples of the fusion results obtained by different feature extractors (convolutional layers, dense blocks and LLRR blocks) in our fusion network.}
	\label{fig:conv-blocks}
\end{figure}

Some examples of the image fusion results are shown in Fig.\ref{fig:conv-blocks}. Compared with the other two architectures, LLRR blocks obtain better fusion performance subjectively. The architecture retains the salient infrared features, irrespective of the number of LLRR blocks used. This shows the merit of the LLRR block in IR-VI image fusion.

As we observed no significant visual difference in the performance of the systems with different number of LLRR blocks (2, 4, 6, 8), the objective evaluations were also conducted to optimize the number of LLRR block in LRRNet. The average values of the six metrics for the different architectures are shown in Table \ref{tab:para-blocks}. 

\begin{table*}[!ht]
	\tiny
	\centering
        \setlength{\abovecaptionskip}{-0.05cm}
	\caption{\label{tab:compare-tno}The average values of the objective metrics obtained by the existing fusion methods and the proposed LRRNet on TNO.}
	\resizebox{0.7\linewidth}{!}{
		\begin{tabular}{c|c |c c c c c c}
			\hline
			&\emph{Year} &En$\uparrow$ &SD$\uparrow$ &MI$\uparrow$ &SSIMm$\uparrow$ & VIFm$\uparrow$ & Nabf$\downarrow$  \\
			\hline
			DenseFuse\cite{li2018densefuse}		&2019 &6.67158 	&67.57282 	&13.34317 	&0.73150 	&0.64576 	&0.09214 \\
			FusionGAN\cite{ma2019fusiongan}		&2019 &6.36285 	&54.35752 	&12.72570 	&0.65384 	&0.45355 	&\emph{\color{red}{0.06706}} \\
			IFCNN\cite{zhang2020ifcnn}			&2020 &6.59545 	&66.87578 	&13.19090 	&0.73186 	&0.59029 	&0.17959 \\
			CUNet\cite{deng2020deep}			&2020 &6.13996 	&43.53543 	&12.27992 	&0.71828 	&0.37522 	&0.16574 \\
			RFN-Nest\cite{li2021rfn}		&2021 &\emph{\color{red}{6.84134}} 	&\emph{\color{red}{71.90131}} 	&\emph{\color{red}{13.68269}} 	&0.72057 	&0.65772 	&0.07288 \\
                Res2Fusion\cite{wang2022res2fusion} &2022 &6.67774   &67.27749      &13.35549 &\emph{\color{red}{0.73216}}     &0.64318 &0.09223 \\
                YDTR\cite{tang2022ydtr}   &2022 &6.22681  &51.48819      &12.45363      &\underline{\textbf{0.75259}}       &0.30924       &\underline{\textbf{0.02167}} \\
                SwinFusion\cite{ma2022swinfusion} &2022 &6.68096   &80.41930      &13.36191 &0.72847     &0.64494 &0.12478 \\
			U2Fusion\cite{xu2020u2fusion}		&2022 &6.75708 	&64.91158 	&13.51416 	&0.70526 	&\underline{\textbf{0.85432}} 	&0.29088 \\
			\hline
			LRRNet			&  &\underline{\textbf{6.85836}} 	&\underline{\textbf{81.78905}} 	&\underline{\textbf{13.71673}} 	&0.70625 	&\emph{\color{red}{0.71399}} 	&0.14168 \\
			\hline
	\end{tabular}}
\end{table*}


When the number of LLRR blocks is set to 4, our fusion network delivers the  best fusion performance (five best values, En, SD, MI, VIFm and Nabf). Thus, in our experiments, the number of LLRR block is set to 4.

\subsubsection{The plot of the loss during the training phase}
\begin{figure}[!ht]
	\centering
	\includegraphics[width=\linewidth]{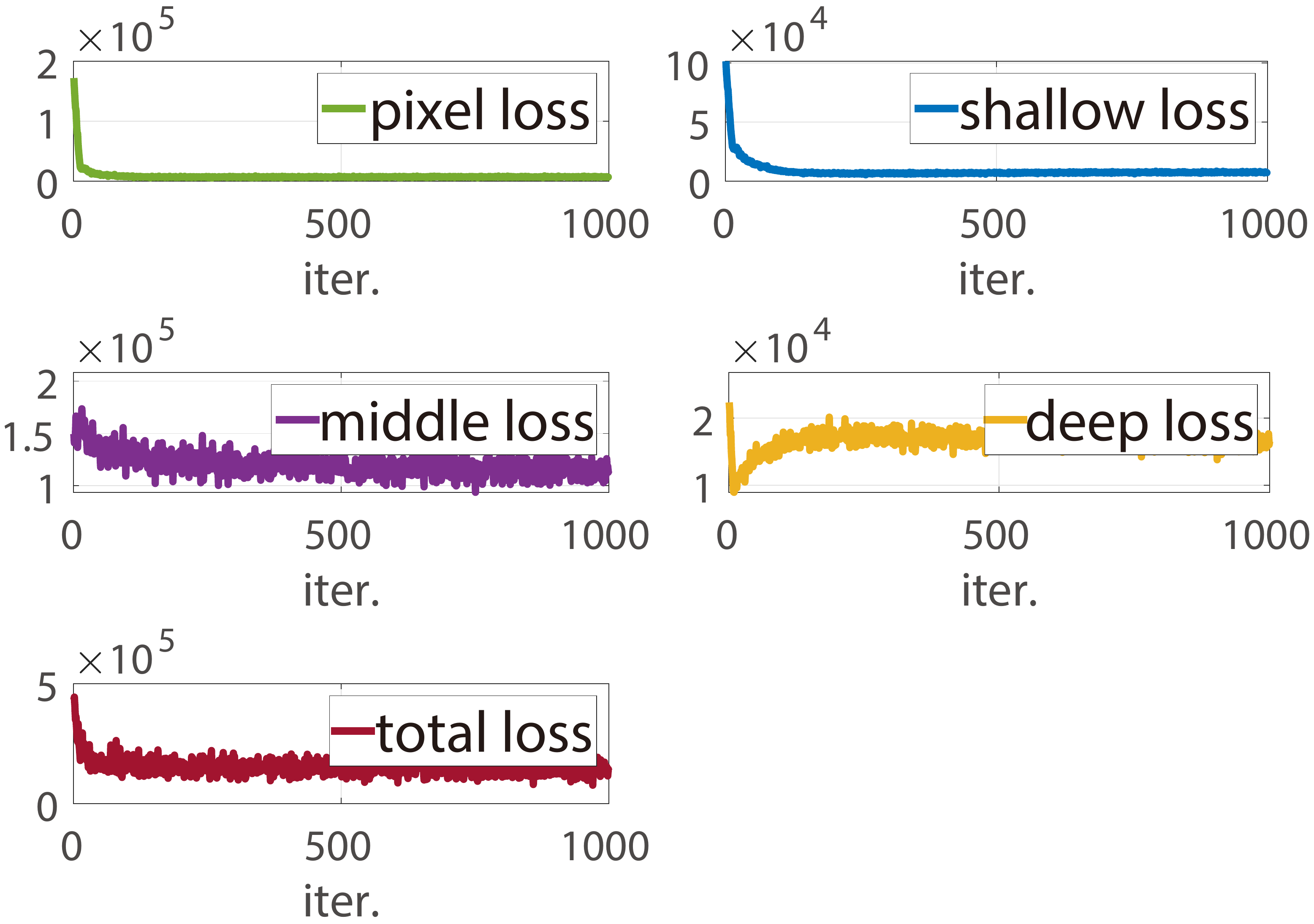}
	\setlength{\abovecaptionskip}{-0.2cm} 
        \setlength{\belowcaptionskip}{-0.2cm} 
	\caption{The loss values for each item of the proposed loss function with the selected optimal hyperparameters. For the lateral axis, each number indicates 10 iterations in training phase.}
	\label{fig:training-plot}
\end{figure}

Thanks to the above ablation studies, we can finally select the optimal hyperparameters of our proposed loss function as $\gamma_1$ = 10, $\gamma_2$ = 1.5, \{$w_{vi}$ = 0.5, $w_{ir}$ = 3.0\}, \{$\beta^2$ = 0.01, $\beta^3$ = 0.5\}, $\gamma_4$ = 2000. The loss values for each component and the total loss are shown in Fig.\ref{fig:training-plot}.

For $L_{pixel}$ and $L_{shallow}$, our proposed model can obtain stable values within a few iterations ($\leq$ 2000), which means it can capture the detail information from both pixel level and shallow feature level very quickly. 

Meanwhile, although the graph of $L_{middle}$ contains some oscillations, it  converges after 5000 iterations. The reason for this phenomenon is that our loss function aims to find the balance between the visible detail information ($L_{pixel}$, $L_{shallow}$) and the infrared salient features ($L_{middle}$).

For $L_{deep}$, in the early stage, the graph exhibits a down-up trend. However, after 2500 iterations, the loss achieves a stable value. The reason is that the deep features (VGG-16, 4-$th$ block), calculated by the $Gram(\cdot)$ operation, contain more abstract information. Hence, the gap between the detail information and the abstract feature becomes larger. Fortunately, with the middle-level features, the proposed loss function can reduce the gap between these two parts. Finally, the $L_{deep}$ also promotes stability, i.e. our network achieves a balance between the preservation of the detail information and the infrared features enhancement.
\subsection{Fusion Results Analysis}

In this section, nine state-of-the-art fusion networks are chosen to preform a comparative evaluation: an auto-encoder based fusion network (DenseFuse)\cite{li2018densefuse}; a GAN based fusion algorithm (FusionGAN)\cite{ma2019fusiongan}; a novel end-to-end fusion method (IFCNN)\cite{zhang2020ifcnn}; an ISTA based fusion a network (CUNet)\cite{deng2020deep}; a novel end-to-end framework based on nest connection and residual fusion network (RFN-Nest)\cite{li2021rfn}; a Res2Net-based fusion network (Res2Fusion)\cite{wang2022res2fusion}; a Y-shape dynamic transformer-based network (YDTR)\cite{tang2022ydtr}; a swin-transformer-based fusion method (SwinFusion)\cite{ma2022swinfusion}; a uniform fusion network (U2Fusion)\cite{xu2020u2fusion}. 
Furthermore, an experiment is also conducted to analyze the number of training parameters for all the comparative networks and our proposed architecture.


%

\begin{table*}[!ht]
       \tiny
       \centering
       \setlength{\abovecaptionskip}{-0.05cm}
       \caption{\label{tab:compare-vot}The average values of the objective metrics obtained by the existing fusion methods and the proposed LRRNet on VOTRGBT-TNO.}
       \resizebox{0.7\linewidth}{!}{
              \begin{tabular}{c|c |c c c c c c }
                     \hline
                     &\emph{Year} & En$\uparrow$ & SD$\uparrow$ & MI$\uparrow$ & SSIMm$\uparrow$ & VIFm$\uparrow$ & Nabf$\downarrow$  \\
                     \hline
                     DenseFuse\cite{li2018densefuse}    &2019 &6.77630      &73.63462     &13.55261     &\emph{\color{red}{0.73282}}       &0.68336      &\emph{\color{red}{0.06346}} \\
                     FusionGAN\cite{ma2019fusiongan}    &2019 &6.52031      &62.84940     &13.04062     &0.67412      &0.48399      &0.07527 \\
                     IFCNN\cite{zhang2020ifcnn}         &2020 &6.74105      &76.24922     &13.48210     &0.72475      &0.74690      &0.20119 \\
                     CUNet\cite{deng2020deep}           &2020 &6.33359      &49.71923     &12.66718     &0.69152      &0.47763      &0.19043 \\
                     RFN-Nest\cite{li2021rfn}    &2021 &6.92952      &78.22247     &13.85904     &0.70572      &0.69406      &0.06357 \\
                Res2Fusion\cite{wang2022res2fusion} &2022 &6.78124     &73.61685      &13.56248 &0.72372     &0.68571 &0.06372 \\
                YDTR\cite{tang2022ydtr}   &2022 &6.40119      &62.44826      &12.80238 &\underline{\textbf{0.73404}}     &0.44251 &\underline{\textbf{0.02648}} \\
                SwinFusion\cite{ma2022swinfusion} &2022 &6.81625  &\underline{\textbf{89.41668}}      &13.63250  &0.70997    &\emph{\color{red}{0.81674}} &0.14224 \\
                     U2Fusion\cite{xu2020u2fusion}      &2022 &\emph{\color{red}{6.94865}}       &76.78378     &\emph{\color{red}{13.89730}}      &0.68804      &\underline{\textbf{0.99065}}      &0.28297 \\
                     \hline
                     LRRNet        & &\underline{\textbf{6.97205}}      &\emph{\color{red}{89.05225}}      &\underline{\textbf{13.94410}}     &0.69851      &0.77659      &0.13162 \\
                     \hline
       \end{tabular}}
\end{table*}

\subsubsection{Fusion Results on TNO}

\begin{figure}[!ht]
	\centering
	\includegraphics[width=0.9\linewidth]{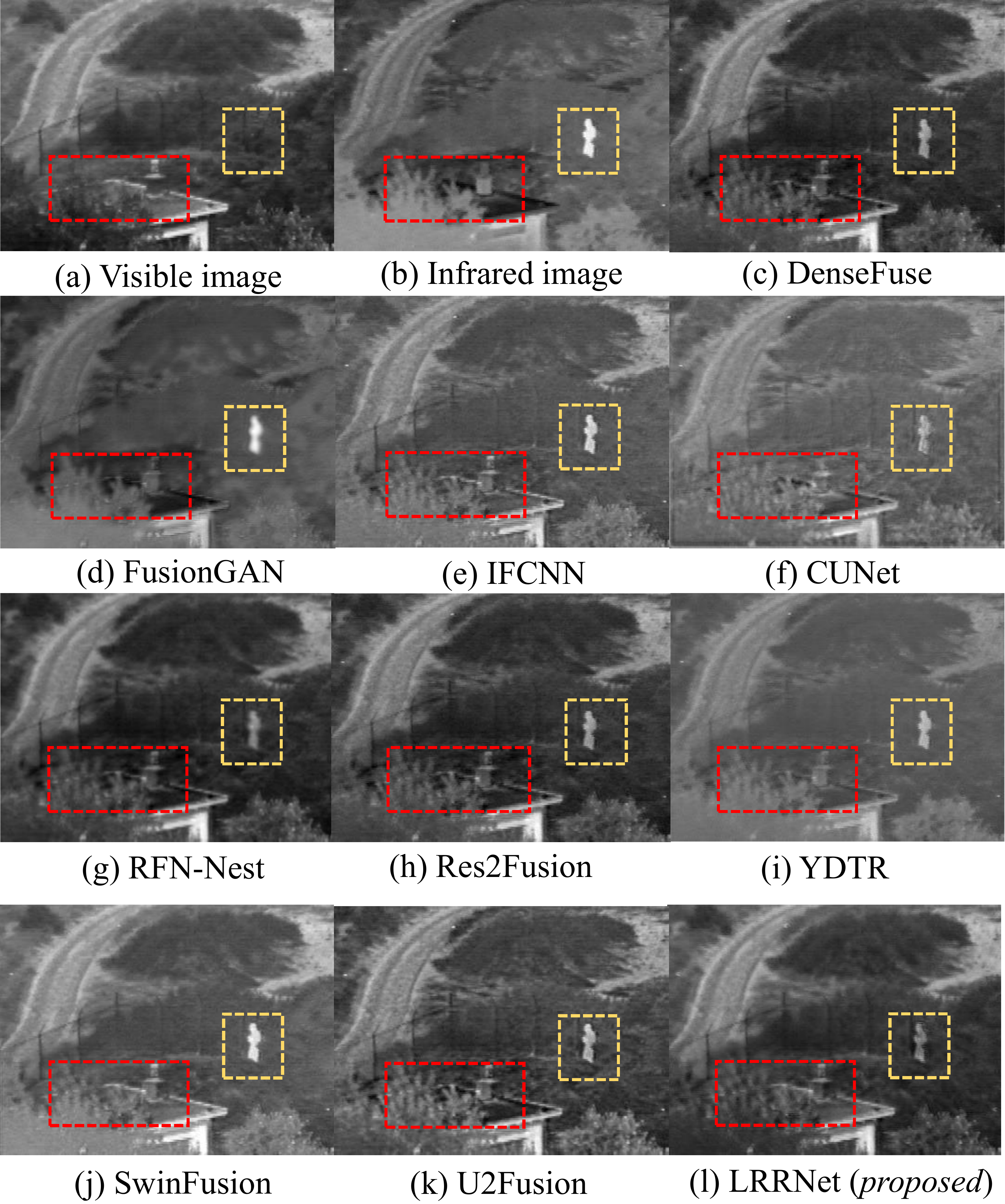}
        \setlength{\abovecaptionskip}{-0.1cm} 
	\caption{The fusion results obtained on the TNO dataset (``man'' image). (a) visible light image; (b) infrared image; (c) DenseFuse; (d) FusionGAN; (e) IFCNN; (f) CUNet; (g) RFN-Nest; (h) Res2Fusion; (i) YDTR; (j) SwinFusion; (k) U2Fusion; (l) LRRNet (proposed).}
	\label{fig:tno-man}
\end{figure}

Firstly, we conduct an experiment on the TNO dataset. The fusion results can be compared subjectively on an example referred to as ``man''. The fused results are shown in Fig.\ref{fig:tno-man}.

Note, the fused images generated by IFCNN, CUNet, YDTR and SwinFusion are blurred, with missing texture details (yellow and red box in Fig.\ref{fig:tno-man}). For DenseFuse, FusionGAN, Res2Fusion and U2Fusion, the textures are preserved but at the same time noise is introduced into the fused images. 

Compared with the existing fusion methods, our proposed LRRNet achieves similar fusion performance in visual assessment. The detail information shown in the visible light image and the infrared salient features of the infrared image are both preserved. Specifically, in Fig.\ref{fig:tno-man}(yellow boxes), although LRRNet does not generate the highlight brightness, the infrared target is shown with better clarity than by the other methods. Note that the textures of target are also preserved. For the tree (red boxes), the textures are also preserved from the visible image and enhanced by the infrared features.

To compare the existing fusion methods with the proposed LRRNet objectively, we use the six metrics discussed earlier. The average values of these metrics are presented in Table \ref{tab:compare-tno}. The best and the second-best values are marked in \underline{\textbf{underline and bold}} and red with italic, respectively.

\begin{figure}[!ht]
	\centering
	\includegraphics[width=0.95\linewidth]{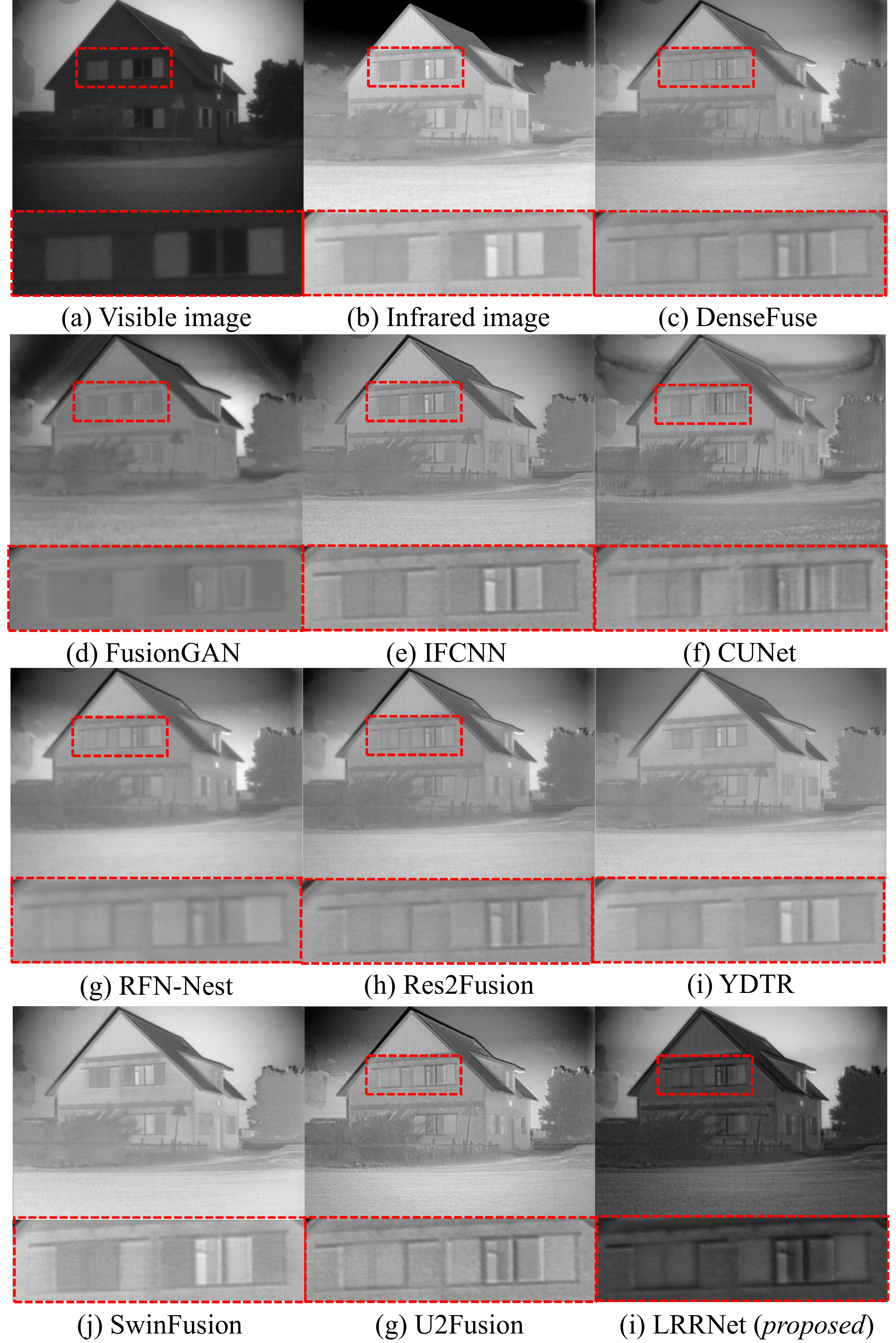}
        \setlength{\abovecaptionskip}{-0.1cm} 
        \caption{The fusion results on VOTRGBT-TNO dataset (``house'' image). (a) visible light image; (b) infrared image; (c) DenseFuse; (d) FusionGAN; (e) IFCNN; (f) CUNet; (g) RFN-Nest; (h) Res2Fusion; (i) YDTR; (j) SwinFusion; (k) U2Fusion; (l) LRRNet (proposed).}
	\label{fig:votrgbt-house}
\end{figure}

\begin{table*}[!ht]
       \centering
       \setlength{\abovecaptionskip}{-0.05cm}
       \caption{\label{tab:num-traing}The number of training parameters for each comparative fusion methods and LRRNet.}
       \resizebox{\linewidth}{!}{
       \begin{tabular}{c|c c c c c c c c c c}
       \hline
       &DenseFuse      &FusionGAN      &IFCNN         &CUNet         &RFN-Nest      &Res2Fusion    &YDTR          &Swinfusion    &U2Fusion      &LRRNet \\
       \emph{Year} &2019\cite{li2018densefuse}      &2019\cite{ma2019fusiongan}      &2020\cite{zhang2020ifcnn}         &2020\cite{deng2020deep}         &2021\cite{li2021rfn}      &2022\cite{wang2022res2fusion}    &2022\cite{tang2022ydtr}          &2022\cite{ma2022swinfusion}    &2022\cite{xu2020u2fusion}      &ours \\
       \hline
       \emph{para.}(M) $\downarrow$   &0.29908        &0.92563        &0.08359       &0.14362       &7.52425       &0.09834       &0.21777       &0.97368       &0.65922       &\underline{\textbf{0.04920}} \\
       \hline
       \end{tabular}}
\end{table*}

Considering all the metrics, the proposed fusion network obtains three best values (En, SD and MI) and one second-best value (VIFm). These suggest the fused images obtained by LRRNet contain more details and less noise. Moreover, the resulting image obtained by our fusion network appears more natural.

\subsubsection{Fusion Results on VOT2020-RGBT}

To analyze the generalization performance of our proposed LRRNet, more IR-VI pairs are selected from TNO \cite{tno2014} and VOT2020-RGBT \cite{vot2020rgbt}.  The fused results of ``house'' image pair are shown in Fig.\ref{fig:votrgbt-house}. 

The results confirm again that the fused images obtained by LRRNet contain more detail information (see red box of Fig.\ref{fig:votrgbt-house}). Moreover, LRRNet produces a more natural image than the other fusion methods.

The results of objective evaluation are presented in Table \ref{tab:compare-vot}. The proposed LRRNet obtains two best values and one second-best value on this expanded IR-VI image fusion dataset, which indicates that LRRNet achieves more robust performance than other state-of-the-art fusion algorithms. Even comparing with the swin-transformer based method SwinFusion, our method can reduce noise (Nabf) while preserve more information (En, MI) from source images.

\subsubsection{The number of training parameters}

\begin{figure}[!ht]
	\centering
	\includegraphics[width=\linewidth]{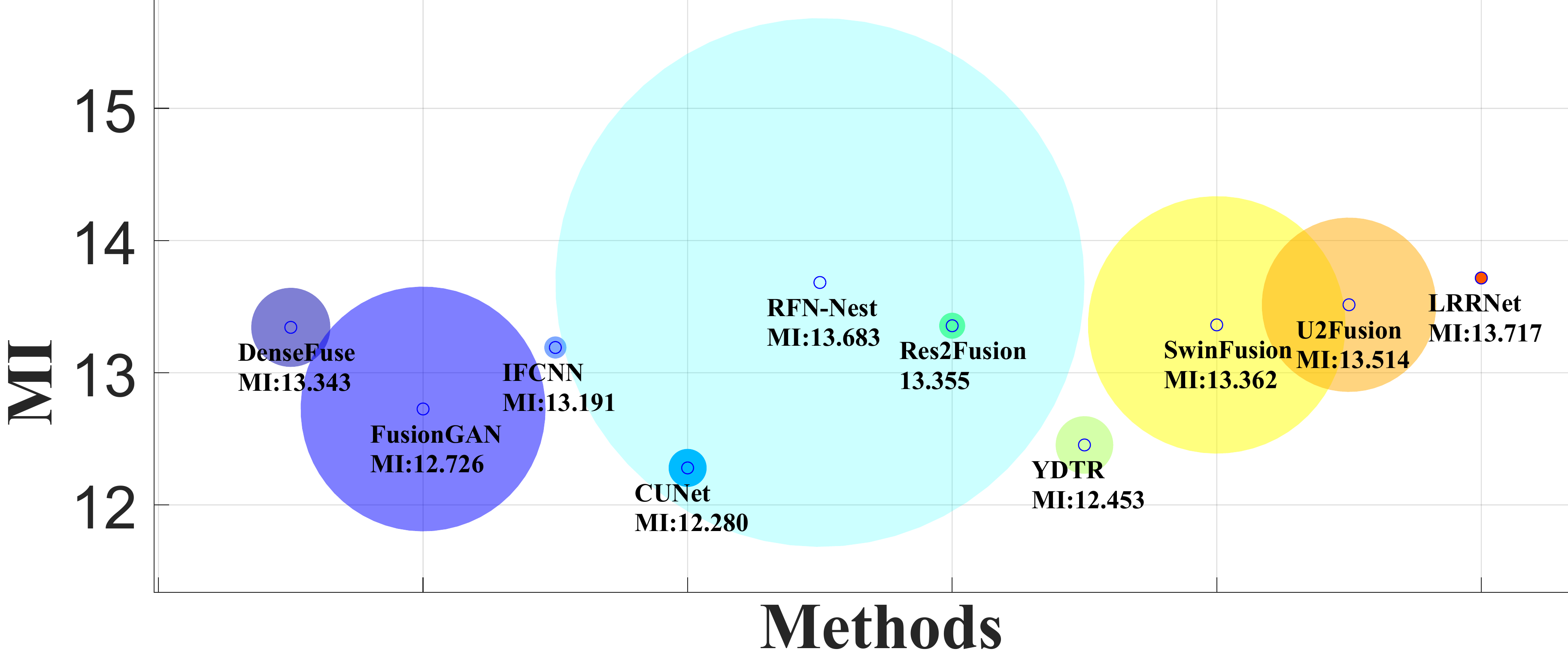}
        \setlength{\abovecaptionskip}{-0.2cm}
	\caption{The number of model parameters for each fusion method and the MI values on TNO dataset. The size of circle indicates the number of training parameters, center point denotes the value of MI.}
	\label{fig:training-para}
\end{figure}

Meanwhile, we also conduct an additional experiment which calculates the number of training parameters for each fusion network, as shown in Table \ref{tab:num-traing} and Fig.\ref{fig:training-para}. For Table \ref{tab:num-traing}, the best values are indicated in \textbf{bold}.

Compared with the lightweight fusion network IFCNN\cite{zhang2020ifcnn}(0.08359M) and Res2Fusion\cite{wang2022res2fusion}(0.09834M),  LRRNet(0.04920M) has the advantage of a smaller network size. Moreover, from Fig.\ref{fig:training-para}, our method (LRRNet) achieves better MI value with a fewer parameters than RFN-Nest (7.52425M). 


\subsection{Experiments on RGBT Object Tracking Task}

To verify the effectiveness of our fusion method (LRRNet) in other computer vision tasks, the RGBT object tracking task is chosen. In the Vision Object Tracking challenge (VOT2020) \cite{vot2020rgbt}, there are 60 videos with both RGB and infrared frames.

In order to apply our fusion framework to the multi-modal object tracking task, a state-of-the-art RGBT tracker (DFAT) \cite{tang2022exploring} is adopted as the baseline tracker. The DFAT won the third place in the evaluation on the public dataset and was the winning tracker in the VOT2020-RGBT challenge.

\begin{figure}[!ht]
	\centering
	\includegraphics[width=0.9\linewidth]{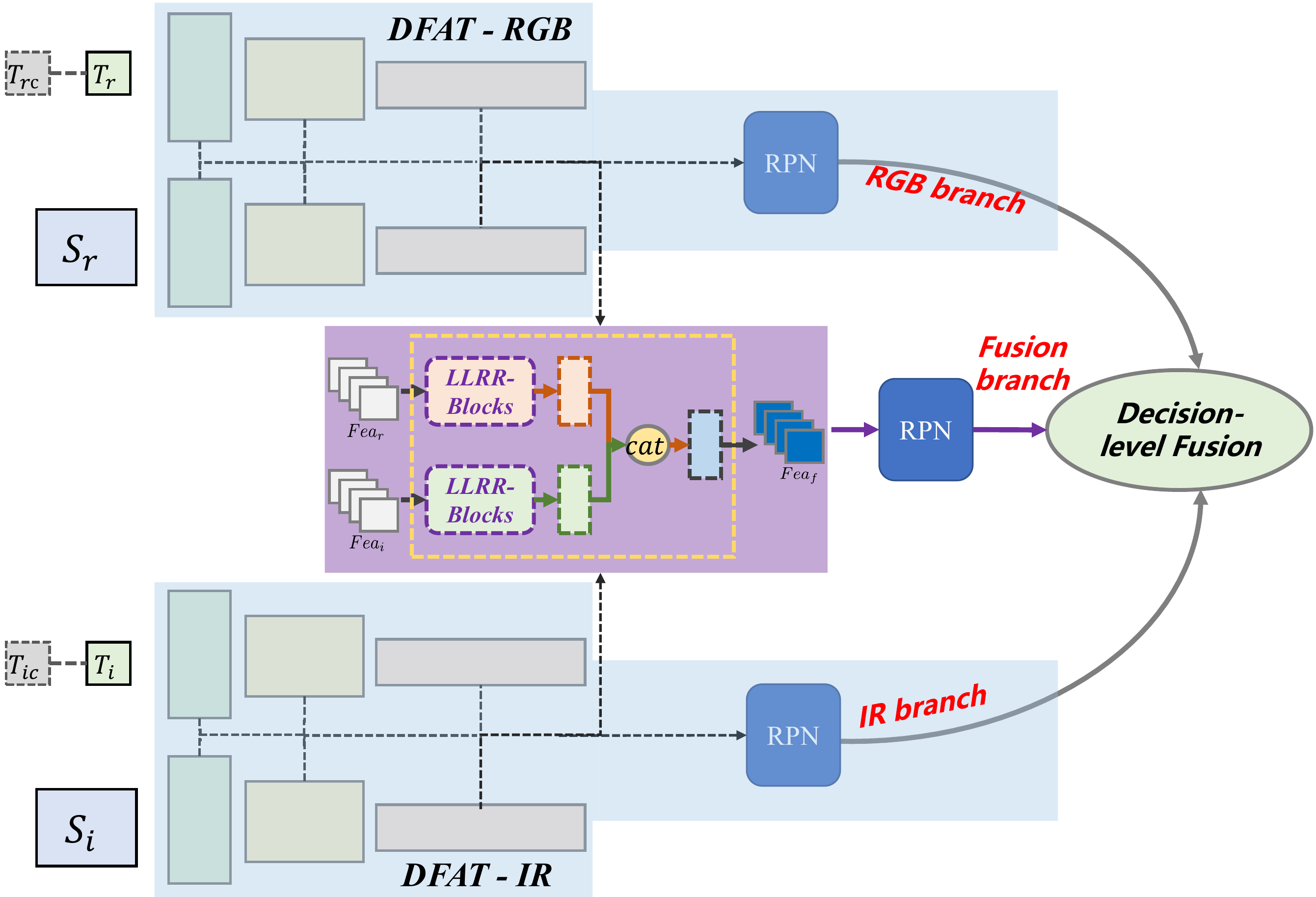}
        \setlength{\abovecaptionskip}{-0.1cm}
	\caption{LRRNet-based RGBT tracker. ``$T_{modal}$'' and ``$S_{modal}$'' denote the template area and the search area for different modalities ($modal \in{\{r, i\}}$), where $r$ and $i$ indicate RGB and infrared, respectively. ``DFAT-RGB'' and ``DFAT-IR'' represent the RGB branch and infrared branch in DFAT. The middle part is the fusion module based on the proposed LLRR-Blocks. ``RPN'' means the region proposal network.}
	\label{fig:tracking-lrrnet}
\end{figure}

In DFAT, a decision-level fusion strategy is applied to fuse the results produced by a single modality tracker(AFAT\cite{xu2020afat}) to achieve a better tracking performance on the RGBT tracking dataset. Our LRRNet module (LLRR blocks) is embedded into DFAT to fuse the RGB and infrared features. Thus, the fusion tracking branch, which is based on LRRNet nodule, is embedded to DFAT\footnote{For more details please refer to supplemental material.}, as shown in Fig.\ref{fig:tracking-lrrnet}.

\subsubsection{The Training Datasets and Loss Function for DFAT}

The training dataset for the RGBT tracking task is insufficient to train an RGBT tracker from scratch. Thus, embedding the multi-modal image (feature) fusion framework into a general tracker is a simple yet efficient solution \cite{li2020mdlatlrr}\cite{li2021rfn}. To train the embedded fusion network (LRRNet module) with DFAT, four limited RGBT datasets are utilized: GTOT \cite{li2016gtot}, VT821\cite{tang2019rgbt}, VT1000\cite{tu2019rgbt}, LasHeR\cite{li2021lasher}.


To fuse the features in DFAT, the additional loss function has three parts: shallow, middle and deep. These three parts correspond to the features extracted by the backbone in DFAT (three outputs in ResNet50). The additional loss function is formulated as follows, 
\begin{eqnarray}\label{equ:loss-track}
	\begin{split}
		Loss_{add}&=\\
		& \gamma_s ||Fea_f^1-ws_{r}Fea_{r}^1||_F^2 \\
		& + \gamma_m ||Fea_f^2-[wm_{i}Fea_{i}^2 + wm_{r}Fea_{r}^2]||_F^2 \\
		& + \gamma_d ||Gram(Fea_f^3) - \\
		& [wd_{i}Gram(Fea_{i}^3) + wd_{r}Gram(Fea_{r}^3)]||_F^2
	\end{split}
\end{eqnarray}
where $Fea_{i}^k$, $Fea_{r}^k$ and $Fea_{f}^k$ ($k=\{1,2,3\}$) indicate the multi-modal inputs ($r$ and $i$ indicate RGB and infrared) and the fused features (output of LRRNet module), respectively. In this experiments, the parameters are set as follows: $\{\gamma_s,\gamma_m,\gamma_d\} = \{10, 2.0, 2000\}$; $ws_{r} = 3.0$; $\{wm_{i}, wm_{r}\} = \{2.0, 1.0\}$; $\{wd_{i}, wd_{r}\} = \{2.0, 1.0\}$.

\subsubsection{The tracking results on VOT2020-RGBT}

\begin{figure}[ht]
	\centering
	\includegraphics[width=0.95\linewidth]{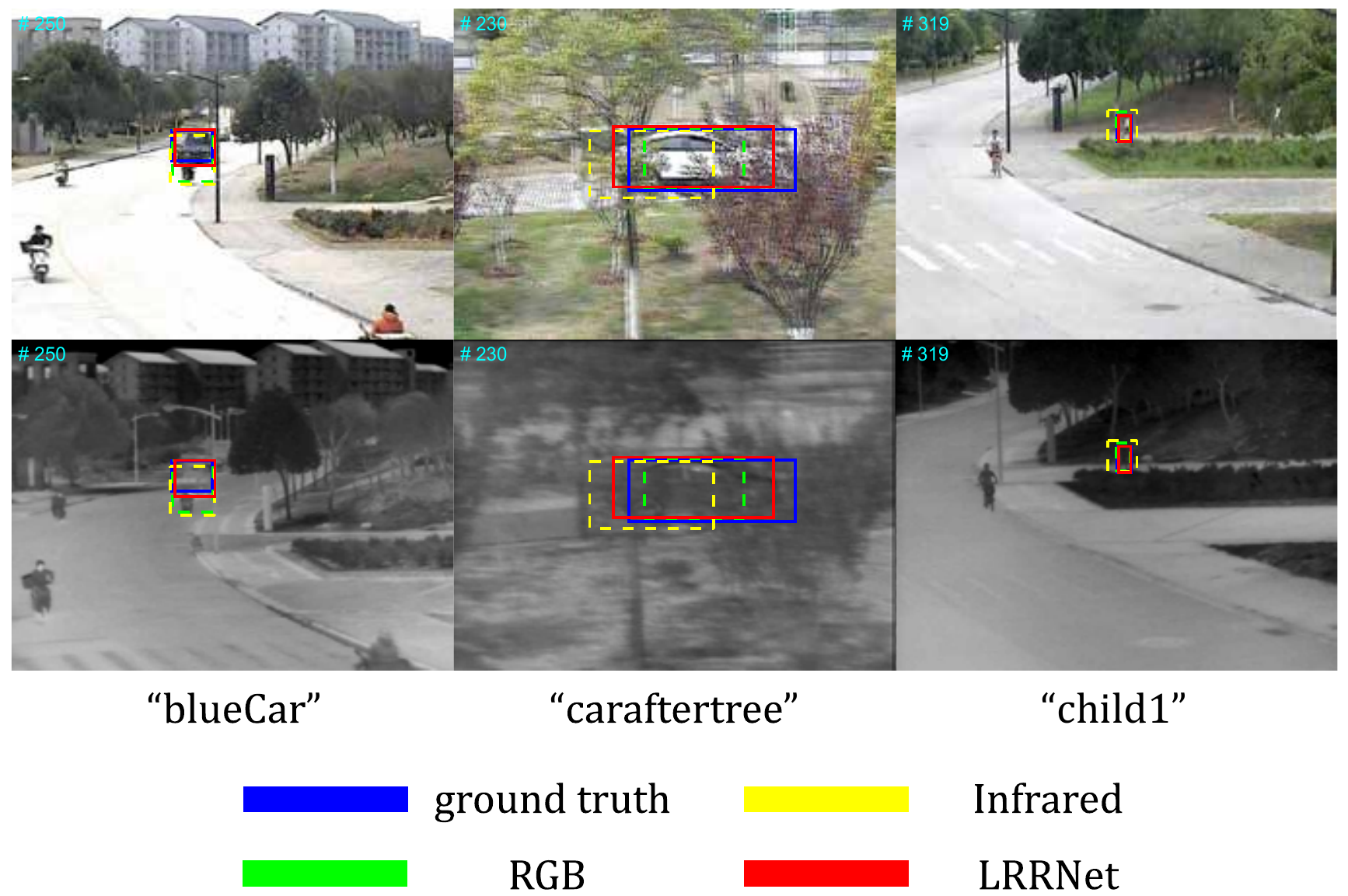}
        \setlength{\abovecaptionskip}{-0.1cm}
	\caption{The tracking results on the VOT2020-RGBT dataset. }
	\label{fig:track}
\end{figure}

In VOT2020\cite{vot2020rgbt}, the adopted evaluation metrics are  $EAO$ (Expected Average Overlap), $A$ (Accuracy) and $R$ (Robustness). The bounding boxes obtained by the single modality tracker (AFAT\cite{xu2020afat}) and the LRRNet-based tracker on several frames are shown in Fig. \ref{fig:track}. 

In the VOT2020-RGBT challenge, the core metric to evaluate the tracking performance is $EAO$. The four trackers used in the comparative experiment are: AFAT\cite{xu2020afat} with single modality (RGB and infrared) as input, combined RFN and AFAT (RFNT)\cite{li2021rfn}, M2C2Frgbt\cite{vot2020rgbt}, and the decision-level fusion tracker (DFAT)\cite{tang2022exploring}. 

Note that AFAT is a RGB tracker. Only one modality (RGB and infrared) is fed into AFAT in this experiment. The RFNT\cite{li2021rfn} is based on AFAT, where the residual fusion network (RFN) is utilized to fuse the multi-modal features and the decision-level fusion is discarded. The M2C2Frgbt is a multi-modal tracker (RGB and infrared) which won the seventh place in the VOT2020-RGBT competition defined on the public dataset. 

To further evaluate the feature fusion performance of ``LRRNet module'', we also chose the recently proposed RFN module \cite{li2021rfn} to create the fusion branch in DFAT, named ``DFAT+RFN''. To train ``DFAT+RFN'', the parameter setting is the same as for ``DFAT+LRRNet'' and the loss function is the same as RFN-Nest\cite{li2021rfn}. The tracking results are shown in Table \ref{tab:tracking2020}.

\begin{table}[!ht]
	\renewcommand\arraystretch{1.1}
	\tiny
	\centering
        \setlength{\abovecaptionskip}{-0.05cm}
	\caption{\label{tab:tracking2020} The tracking results (evaluation values) obtained on the VOT2020-RGBT dataset.}
	\resizebox{0.95\linewidth}{!}{
	\begin{tabular}{c|c|c c c}
		\hline
		\multicolumn{2}{c|}{\emph{VOT2020}} &$EAO$ &$A$ &$R$ \\
		\hline
		\multirow{2}*{AFAT\cite{xu2020afat}} &    
		\emph{RGB}             &0.3291      &0.6352        &0.6691 \\
		&\emph{Infrared}  &0.2653    &0.5733        &0.5881 \\
		\hline		
		\multicolumn{2}{c|}{RFNT\cite{li2021rfn}(2019)}		  &0.3710 	&0.6680 	&0.7260\\
		\multicolumn{2}{c|}{M2C2Frgbt\cite{vot2020rgbt}(2020)}	&0.3320 	&0.6360 	&0.7220 \\
		\multicolumn{2}{c|}{DFAT\cite{tang2022exploring}(2020, \emph{winner})}	&\emph{\color{red}{0.4079}} 	&\emph{\color{red}{0.6739}} 	&\emph{\color{red}{0.7785}} \\
		\multicolumn{2}{c|}{DFAT+RFN\cite{li2021rfn}(2021)}		  &0.3840 	&0.6522 	&0.7607 \\
		\multicolumn{2}{c|}{DFAT+LRRNet (\emph{ours})}  &\underline{\textbf{0.4143}} 	&\underline{\textbf{0.6780}} 	&\underline{\textbf{0.7929}} \\
		\hline
	\end{tabular}}
\end{table}

Compared with the single modality tracker (AFAT with RGB or infrared), all the multi-modal trackers achieve better performance. Moreover, the LRRNet-based tracker (``DFAT+LRRNet'') achieves the best tracking performance compared with the other RGBT trackers (even the winner tracker DFAT) according to the $EAO$ metric. 

The only difference between  ``DFAT+LRRNet'' and ``DFAT+RFN'' is that the the multi-modal feature fusion network (RFN vs LRRNet) in the fusion branch. The best values of the three measures indicate that the proposed LRRNet exhibits better fusion performance than RFN\cite{li2021rfn} in both the image fusion task and the RGBT tracking task. 

The main reasons why ``DFAT+RFN'' does not achieve better performance is that the network structure is too simple to use with the decision-level based tracker, and the loss function of RFN also needs to be reconsidered. In contrast, these two drawbacks are addressed in our proposed LRRNet module.

This experiment indicates that the proposed LRRNet is not only effective in multi-modal image fusion tasks, but also in multi-modal computer vision tasks, such as RGBT tracking. With LRRNet, the tracker performance is considerably improved.

\section{Conclusion}

In this paper, to exploit a novel network design scheme and to avoid the trial and error strategy in the construction of a fusion network, a fusion task driven optimal model is introduced and becomes the essence of the proposed end-to-end fusion network architecture. Due to its powerful feature representation ability, LRR is introduced into our model. LRR based decomposition model is designed for the fusion task in which the source image is decomposed into base features and detail features. Then, a learning algorithm is utilized to optimize the proposed fusion task driven model (LLRR) and to construct a novel fusion network architecture. Thus, different from the empirically designed network, our proposed lightweight network architecture is defined by a specific task model. Furthermore, a novel multi-level (from detail to semantic information) loss function is proposed to train the LLRR block based end-to-end fusion network (LRRNet).

Compared with nine classical and state-of-the-art fusion methods, in two public IR-VI image datasets, LRRNet obtains sharper outputs.
It preserves more detail information from the visible image and enhances the infrared features. Moreover, the network size of LRRNet is significantly more lightweight than the model size of the other methods.

We also demonstrated the merit of the proposed feature fusion network (LRRNet module) in other computer vision tasks (such as object tracking). These experiments illustrate that our LRRNet is an efficient network for IR-VI image fusion which can be well applied to other computer vision tasks to improve the algorithm performance. Although we demonstrated the merits of LRRNet on a single  application concerned with RGBT tracking, we believe the proposed model is applicable to other computer vision tasks. In future, we intend to explore other application directions for multi-modal image fusion.

\ifCLASSOPTIONcompsoc
  \section*{Acknowledgments}
\else
  \section*{Acknowledgment}
\fi

This work was supported by the National Natural Science Foundation of China (62020106012, 62202205, U1836218, 62106089, 62006097), the Fundamental Research Funds for the Central Universities (JUSRP123030), the 111 Project of Ministry of Education of China (B12018), the Natural Science Foundation of Jiangsu Province(Grant No.BK20200593), and the Engineering and Physical Sciences Research Council (EPSRC) (EP/N007743/1, MURI/EPSRC/DSTL, EP/R018456/1).

\ifCLASSOPTIONcaptionsoff
  \newpage
\fi



\bibliographystyle{IEEEtran}
\bibliography{bibtex.bib}
%
%
%

%

\begin{IEEEbiography}
[{\includegraphics[width=1in,height=1.25in,clip,keepaspectratio]{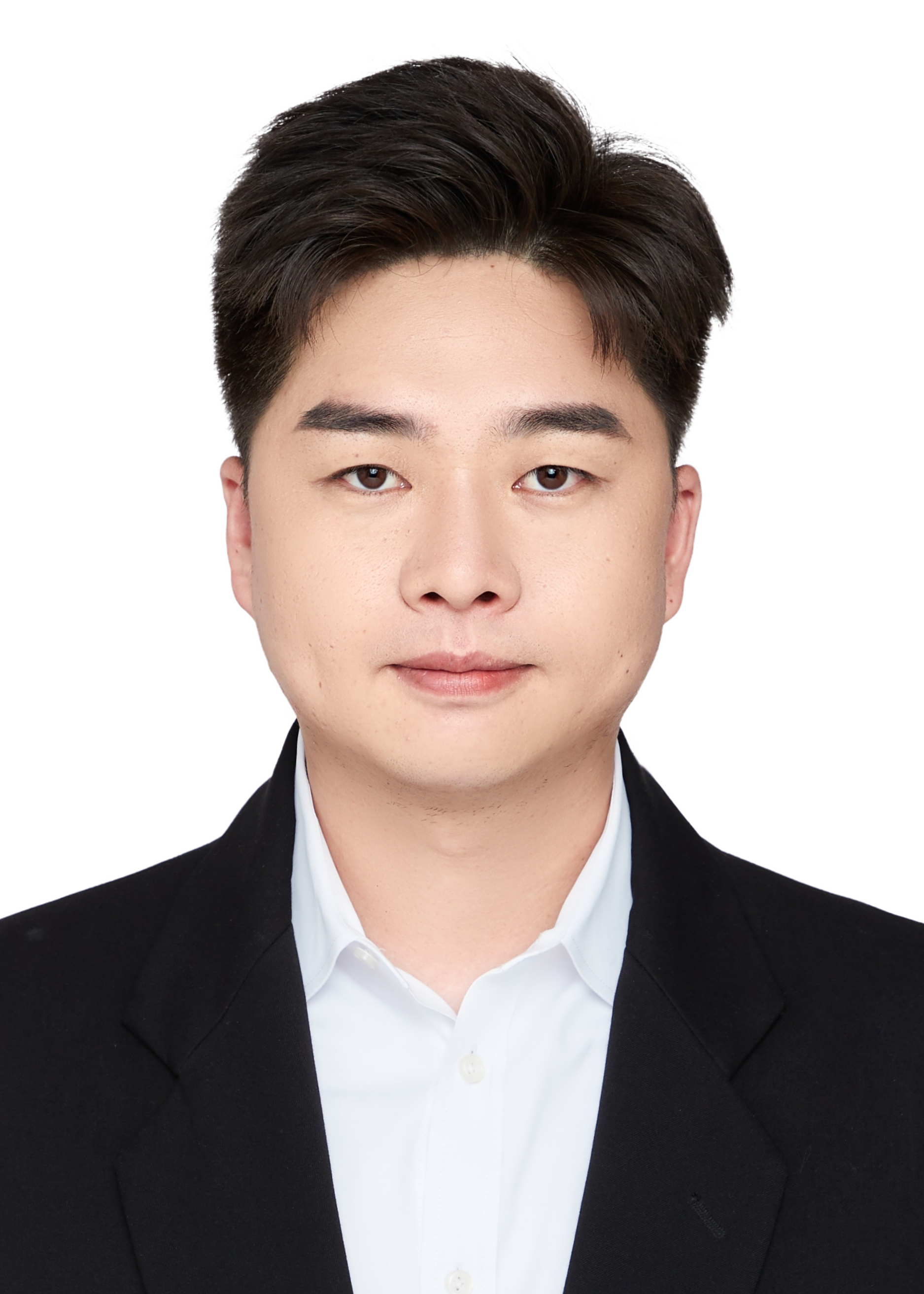}}]{Hui Li}
received the B.Sc. degree in School of Internet of Things Engineering from Jiangnan University, China, in 2015. He received the PhD degree at the School of Internet of Things Engineering, Jiangnan University, Wuxi, China, in 2022. He is currently a Lecturer at the School of Artificial Intelligence and Computer Science, Jiangnan University, Wuxi, China. His research interests include image fusion and multi-modal visual information processing. He has been chosen among the World's Top 2\% Scientists ranking in the single recent year dataset published by Stanford University, Version 5, Nov 2022.

He has published several scientific papers, including Information Fusion, IEEE TIP, IEEE TCYB, IEEE TIM, ICPR etc. He achieved top tracking performance in several competitions, including the VOT2020 RGBT challenge (ECCV20) and Anti-UAV challenge (ICCV21).

\end{IEEEbiography}

\begin{IEEEbiography}[{\includegraphics[width=1in,height=1.25in,clip,keepaspectratio]{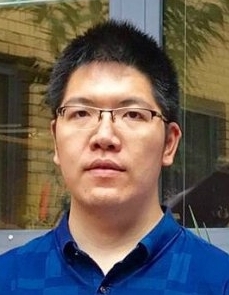}}]{Tianyang Xu} 
received the B.Sc. degree in electronic science and engineering from Nanjing University, Nanjing, China, in 2011. He received the PhD degree at the School of Artificial Intelligence and Computer Science, Jiangnan University, Wuxi, China, in 2019. He was a research fellow at the Centre for Vision, Speech and Signal Processing (CVSSP), University of Surrey, Guildford, United Kingdom, from 2019 to 2021. He is currently an Associate Professor at the School of Artificial Intelligence and Computer Science, Jiangnan University, Wuxi, China. His research interests include visual tracking and deep learning. 
	
He has published several scientific papers, including IJCV, ICCV, TIP, TIFS, TKDE, TMM, TCSVT etc. He achieved top 1 tracking performance in several competitions, including the VOT2018 public dataset (ECCV18), VOT2020 RGBT challenge (ECCV20), and Anti-UAV challenge (CVPR20).
\end{IEEEbiography}

\begin{IEEEbiography}[{\includegraphics[width=1in,height=1.25in,clip,keepaspectratio]{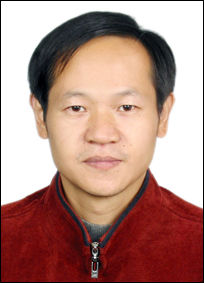}}]{Xiao-Jun Wu}
received his B.S. degree in mathematics from Nanjing Normal University, Nanjing, PR China in 1991 and M.S. degree in 1996, and Ph.D. degree in Pattern Recognition and Intelligent System in 2002, both from Nanjing University of Science and Technology, Nanjing, PR China, respectively. He was a fellow of United Nations University, International Institute for Software Technology (UNU/IIST) from 1999 to 2000. From 1996 to 2006, he taught in the School of Electronics and Information, Jiangsu University of Science and Technology where he was an exceptionally promoted professor. He joined Jiangnan University in 2006 where he is currently a distinguished professor in the School of Artificial Intelligence and Computer Science, Jiangnan University. 

He won the most outstanding postgraduate award by Nanjing University of Science and Technology. He has published more than 400 papers in his fields of research. He was a visiting postdoctoral researcher in the Centre for Vision, Speech, and Signal Processing (CVSSP), University of Surrey, UK from 2003 to 2004, under the supervision of Professor Josef Kittler. His current research interests are pattern recognition, computer vision, fuzzy systems, and neural networks. He owned several domestic and international awards because of his research achievements. Currently, he is a Fellow of IAPR and AAIA. 
\end{IEEEbiography}

\begin{IEEEbiography}[{\includegraphics[width=1in,height=1.25in,clip,keepaspectratio]{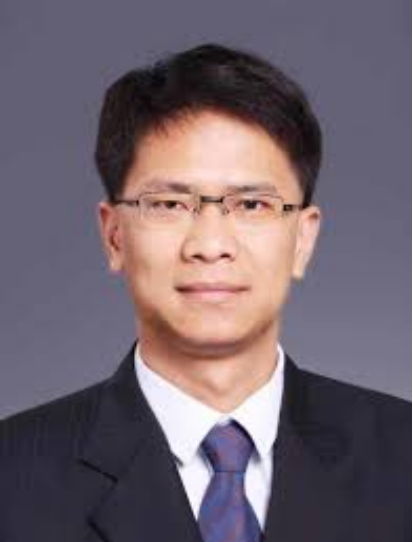}}]{Jiwen Lu} 
(Senior Member, IEEE) received the BEng degree in mechanical engineering and the MEng degree in electrical engineering from the Xian University of Technology, Xian, China, in 2003 and 2006, respectively, and the PhD degree in electrical engineering from Nanyang Technological University, Singapore, in 2012. He is currently an associate professor with the Department of Automation, Tsinghua University, Beijing, China. His current research interests include computer vision and pattern recognition. 
    
He was/is a member of the Image, Video and Multidimensional Signal Processing Technical Committee, Multimedia Signal Processing Technical Committee and the Information Forensics and Security Technical Committee of the IEEE Signal Processing Society, a member of the Multimedia Systems and Applications Technical Committee and the Visual Signal Processing and Communications Technical Committee of the IEEE Circuits and Systems Society, respectively. He serves as the General Co-Chair for the International Conference on Multimedia and Expo (ICME) 2022, the Program Co-Chair for the International Conference on Multimedia and Expo 2020, the International Conference on Automatic Face and Gesture Recognition (FG) 2023, and the International Conference on Visual Communication and Image Processing (VCIP) 2022. He serves as the co-editor-of-chief for Pattern Recognition Letters, an associate editor for IEEE Transactions on Image Processing, IEEE Transactions on Circuits and Systems for Video Technology, IEEE Transactions on Biometrics, Behavior, and Identity Sciences, and Pattern Recognition. He was a recipient of the National Natural Science Funds for Distinguished Young Scholar. He is a fellow of IAPR.
\end{IEEEbiography}

\begin{IEEEbiography}[{\includegraphics[width=1in,height=1.25in,clip,keepaspectratio]{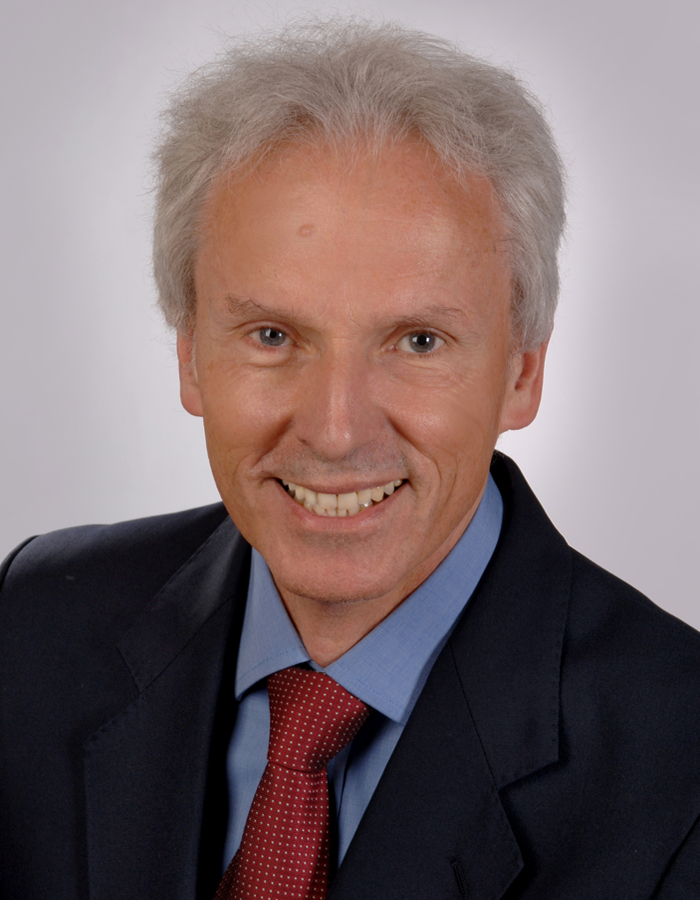}}]{Josef Kittler} 
(M’74-LM’12) received the B.A.,	Ph.D., and D.Sc. degrees from the University of Cambridge, in 1971, 1974, and 1991, respectively. He is a distinguished Professor of Machine Intelligence	at the Center for Vision, Speech and Signal Processing, University of Surrey, Guildford, U.K. He conducts research in biometrics, video and image	database retrieval, medical image analysis, and	cognitive vision. He published the textbook Pattern Recognition: A Statistical Approach and over 700	scientific papers. His publications have been cited	more than 60,000 times (Google Scholar).
    
He is series editor of Springer Lecture Notes on Computer Science. He currently serves on the Editorial Boards of Pattern Recognition Letters, Pattern Recognition and Artificial Intelligence, Pattern Analysis and Applications. He	also served as a member of the Editorial Board of IEEE Transactions on	Pattern Analysis and Machine Intelligence during 1982-1985. He served on	the Governing Board of the International Association for Pattern Recognition	(IAPR) as one of the two British representatives during the period 1982-2005, President of the IAPR during 1994-1996.
\end{IEEEbiography}

\end{document}